\documentclass{article}

\usepackage{arxiv}

\usepackage[utf8]{inputenc} 
\usepackage[T1]{fontenc}    
\usepackage{hyperref}       
\usepackage{url}            
\usepackage{booktabs}       
\usepackage{amsfonts}       
\usepackage{nicefrac}       
\usepackage{microtype}      
\usepackage{lipsum}		
\usepackage{microtype}
\usepackage{graphicx}
\usepackage{subfigure}
\usepackage{booktabs} 
\usepackage{amsmath}
\usepackage{amssymb}
\usepackage{natbib}

\def\floor#1{\lfloor #1 \rfloor}
\usepackage{hyperref}
\usepackage{bm}
\usepackage{xr}
\usepackage{diagbox}
\RequirePackage{fancyhdr}
\RequirePackage{color}
\RequirePackage{algorithm}
\RequirePackage{algorithmic}
\RequirePackage{natbib}
\RequirePackage{eso-pic} 
\RequirePackage{forloop}
\newcommand{\Mean}{{\mathbb{E}}}

\newcommand{\prob}{{\mathbb{P}}}
\DeclareMathOperator*{\argmin}{arg\,min}
\DeclareMathOperator*{\argmax}{arg\,max}
\newtheorem{thm}{Theorem}
\newtheorem{lemma}{Lemma}
\newcommand\independent{\protect\mathpalette{\protect\independenT}{\perp}}
\def\independenT#1#2{\mathrel{\rlap{$#1#2$}\mkern2mu{#1#2}}}
\graphicspath{{figs0/}}

\title{Does the Markov Decision Process Fit the Data: Testing for the Markov Property in Sequential Decision Making}

\author{
	Chengchun Shi\\
	London School of Economics and Political Science
   \And
  Runzhe Wan \\
  North Carolina State University
  \AND
  Rui Song\\
  North Carolina State University
  \AND
  Wenbin Lu\\
  North Carolina State University
  \AND
  Ling Leng\\
  Amazon
}
\date{}

\begin{document}
\maketitle

\begin{abstract}
	The Markov assumption (MA) is fundamental to the empirical validity of reinforcement learning. 
	In this paper, we propose a novel Forward-Backward Learning procedure to test MA in sequential decision making. The proposed test does not assume any parametric form on the joint distribution of the observed data and plays an important role for identifying the optimal policy in high-order Markov decision processes and partially observable MDPs. We apply our test to both synthetic datasets and a real data example from mobile health studies to illustrate its usefulness. 
\end{abstract}

\section{Introduction}
\label{secintroduction}
Reinforcement learning (RL) is a general technique that allows an agent to learn and interact with an environment. In RL, the state-action-reward triplet is typically modelled by the Markov decision process \citep[MDP, see e.g.][]{Puterman1994}. Central to the empirical validity of various RL algorithms is the Markov assumption (MA). Under MA, there exists an optimal stationary policy that is no worse than any non-stationary or history dependent policies \citep{Puterman1994,Sutton2018}. When this assumption is violated, the optimal policy might depend on lagged variables and any stationary policy can be sub-optimal. Thus, MA forms the basis for us to select the set of state variables to implement RL algorithms. 
The focus of this paper is to test MA in sequential decision making problems. 

\subsection{Contributions and advances of our test}
First, our test is useful in identifying the optimal policy in high-order MDPs (HMDPs). Under HMDPs, the optimal policy at time $t$  depends not only on the current covariates $S_{0,t}$, but also the past state-action pairs  $(S_{0,t-1},A_{0,t-1})$, $\cdots$,  $(S_{0,t-\kappa_0+1},A_{0,t-\kappa_0+1})$ for some $\kappa_0>1$ (see Lemma \ref{lemma2} for a formal statement). In real-world applications, it remains challenging to properly select the look-back period $\kappa_0$. On one hand, $\kappa_0$ shall be sufficiently large to guarantee MA holds. On the other hand, including too many lagged variables will result in a very noisy policy. To determine $\kappa_0$, we propose to construct the state by concatenating measurements taken at time points $t,\cdots,t-k+1$ and sequentially apply our test for $k=1,2,\cdots,$ until the null hypothesis MA is not rejected. Then we use existing RL algorithms based on the constructed state to estimate the optimal policy. We apply such a procedure to both synthetic and real datasets in Section \ref{sec:ohio}. Results show that the estimated policy based on our constructed states achieves the largest value in almost all cases. 

Second, our test is useful in detecting partially observable MDPs. 
Suppose we 
concatenate measurements over sufficiently many decision points and our test still rejects MA.  
Then we shall consider modelling the system dynamics by partially observable MDPs (POMDPs) or other non-Markovian problems. Applying RL algorithms designed for these settings have been shown to yield larger value functions than those for standard MDPs \citep[see e.g.][]{hausknecht2015}. In Section \ref{sec:tiger}, we 
illustrate the usefulness of our test in detecting POMDPs. 

Third, we propose a novel testing procedure to test MA. To the best of our knowledge, this is the first work on developing valid statistical tests for MA in sequential decision making. Major challenges arise when the state vector is high-dimensional. This is certainly the case as we convert the process into an MDP by concatenating data over multiple decision points. To deal with high-dimensionality, we proposed a novel forward-backward learning procedure to construct the  test statistic. The key ingredient of our test lies in constructing a doubly robust estimating equation to alleviate biases of modern machine learning algorithms. This ensures our test statistic has a tractable limiting distribution. 
In addition, since the test is constructed based on forward and backward learners (see Section \ref{secFBML} for details) estimated using the state-of-the-art machine learning estimation methods, it is well-suited to  high-dimensional settings. 

Lastly, 
our test is valid as either the number of trajectories $n$ or the number of decision points $T$ in each trajectory diverges to infinity. It can thus be applied to a variety of sequential decision making problems ranging from the Framingham heart study \citep{tsao2015} with over two thousand trajectories to the OhioT1DM dataset \citep{marling2018} that contains eight weeks' worth of data for six trajectories. Our test can also be applied to applications from video games where both $n$ and $T$ approach infinity. 



\subsection{Related work}\label{secrelatedwork}
There exists a huge literature on developing RL algorithms. Some recent popular methods include fitted Q-iteration \citep{riedmiller2005}, deep Q-network \citep{mnih2015}, double Q-learning \citep{van2016}, asynchronous advantage actor-critic \citep{mnih2016}, etc. All the above mentioned methods model the sequential decision making problems by MDPs. When the Markov assumption is violated, the foundation of these algorithms is shaking hence may lead to deterioration of their performance to different degrees. 

Currently, only a few methods have been proposed to test the Markov assumption. Among those available, \citet{Chen2012} developed such a test in time series analysis. Constructing their test statistic requires to estimate the conditional characteristic function (CCF) of the current measurements 
given those taken in the past. 
\citet{Chen2012} proposed to estimate the CCF based on local polynomial regression \citep{stone1977}. 
We note their method cannot be directly used to test MA in MDP. Even though we can extend their method to our setup, the resulting test will perform poorly in settings where the dimension of the state vector is large, since local polynomial fitting suffers from the curse of dimensionality.  

Our work is also related to the literature on conditional independence testing \citep[see e.g.][]{zhang2012kernel,Su2014,Wang2015,Huang2016,Wang2018,berrett2020}. However, all the above methods require observations to be independent and are not suitable to our settings where measurements are time dependent.   

\subsection{Organization of the paper}
The rest of the paper is organized as follows. In Section \ref{secformulation}, we introduce the MDP, HMDP and POMDP models, and establish the existence of the optimal stationary policy under MA. In Section \ref{sectest}, we introduce our testing procedure for MA and prove the validity of our test. In Section \ref{modelselect}, we introduce a forward procedure based on our test for model selection. Empirical studies are presented in Section \ref{secnumerical}. 

\section{Model setup}\label{secformulation}
\subsection{MDP and existence of the optimal stationary policy}\label{secMDP}
Let $(S_{0,t},A_{0,t},R_{0,t})$ denote the state-action-reward triplet collected at time $t$. For any integer $t\ge 0$, let $\bar{\bm{S}}_{0,t}=(S_{0,0},A_{0,0},S_{0,1},A_{0,1},\cdots,S_{0,t})^\top$ denote the state and action history. For simplicity, we assume the action set $\mathcal{A}$ is finite and the rewards are uniformly bounded. In MDPs, it is typically assumed that the following Markov assumption holds, 
\begin{figure}
	\centering
	\includegraphics[width=9cm]{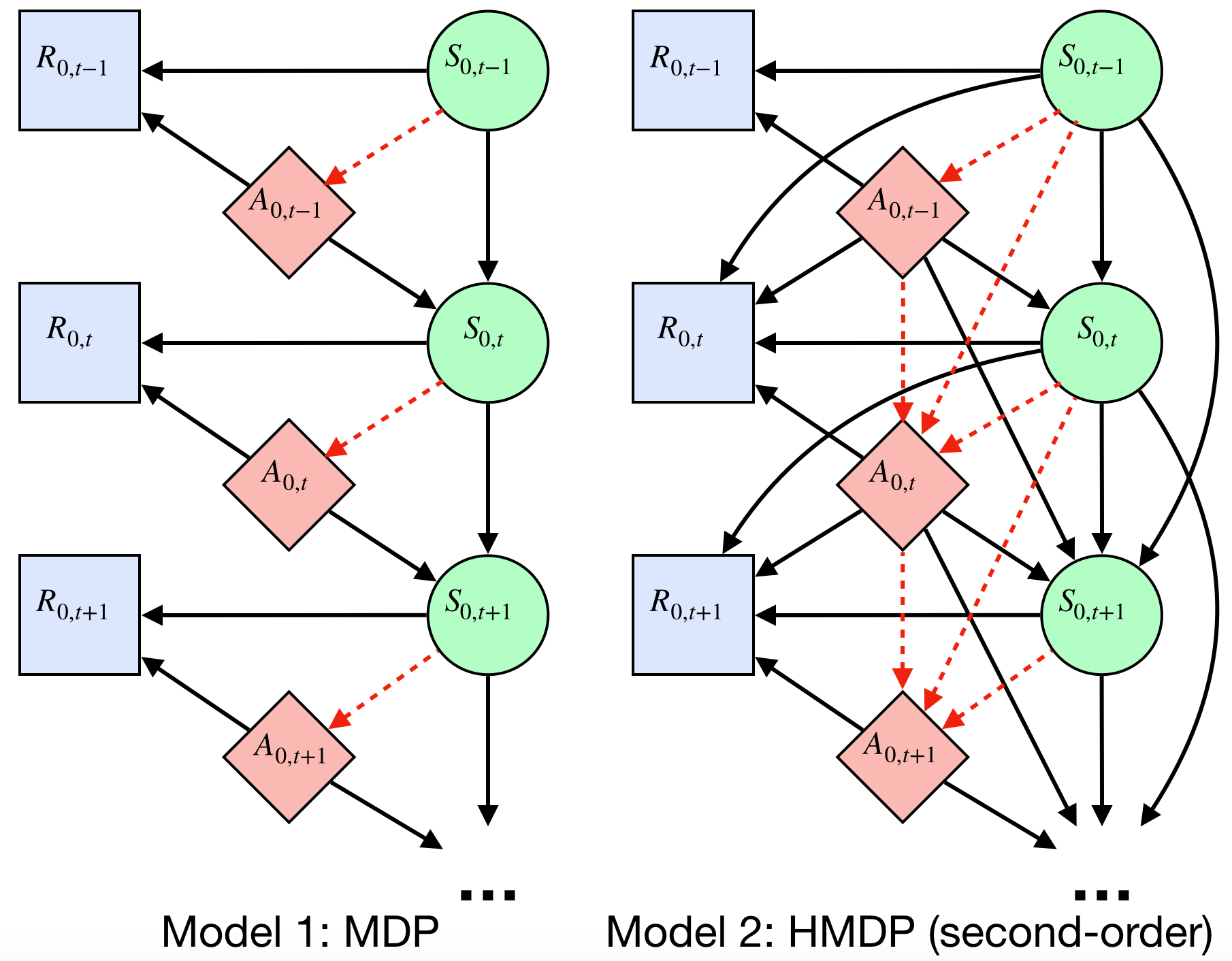}
	\includegraphics[width=5.25cm]{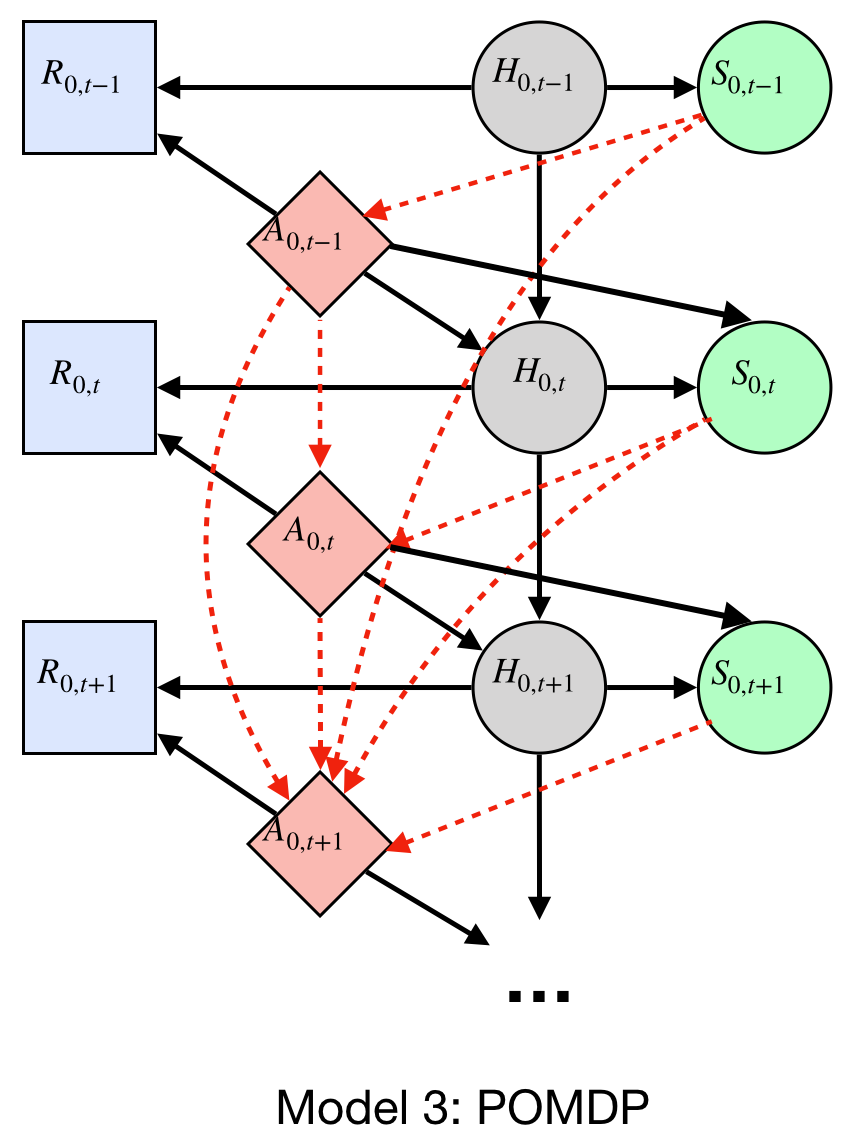}
	\caption{Causal diagrams for MDPs, HMDPs and POMDPs. The solid lines represent the causal relationships and the dashed lines indicate the information needed to implement the optimal policy.}\label{fig1}
\end{figure}
\begin{eqnarray*}
	\prob (S_{0,t+1}\in \mathcal{S},R_{0,t}\in \mathcal{R}|A_{0,t},\bar{\bm{S}}_{0,t}, \{R_{0,j}\}_{j<t} )\\
	=\mathcal{P}( \mathcal{S},\mathcal{R};A_{0,t},S_{0,t}),
\end{eqnarray*}
for some Markov transition kernel $\mathcal{P}$ and any $\mathcal{S}\subseteq \mathbb{S}$, $\mathcal{R}\subseteq \mathbb{R}$, $t\ge 0$ where $\mathbb{S}\in \mathbb{R}^p$ denotes the state space. 

A \textit{history-dependent} policy $\pi$ is a sequence of decision rules $\{\pi_t\}_{t\ge 0}$ where each $\pi_t$ maps $\bar{\bm{S}}_{0,t}$ to a probability mass function $\pi_t(\cdot|\bar{\bm{S}}_{0,t})$ on $\mathcal{A}$. 
When there exists some function $\pi^*$ such that $\pi_t(\cdot|\bar{\bm{S}}_{0,t})=\pi^*(\cdot|S_{0,t})$ for any $t\ge 0$ almost surely, we refer to $\pi$ as a \textit{stationary} policy. 

For a given discounted factor $0<\gamma<1$, the objective of RL is to learn an optimal policy $\pi=\{\pi_t\}_{t\ge 0}$ that maximizes the value function
\begin{eqnarray*}
	V(\pi;s)=\sum_{t=0}^{+\infty} \gamma^t \Mean^{\pi_t} (R_{0,t}|S_{0,0}=s),
\end{eqnarray*}
for any $s\in \mathbb{S}$, where the expectation $\Mean^{\pi_t}$ is taken by assuming that the system follows $\pi_t$. Let HR and SR denote the class of history-dependent and stationary policies, respectively. The following lemma forms the basis of existing RL algorithms.

\begin{lemma}\label{keylemma}
	Under MA, there exists some $\pi^{\tiny{opt}}\in \hbox{SR}$ such that 
	$V(\pi^{\tiny{opt}};s)=\sup_{\pi\in \hbox{\scriptsize{HR}}} V(\pi;s)$ for any $s\in \mathbb{S}$. 
\end{lemma}

Lemma \ref{keylemma} implies that under MA, it suffices to restrict attention to stationary policies. This greatly simplifies the estimating procedure of the optimal policy. When MA is violated however, we need to focus on history-dependent policies as they may yield larger value functions.

When the state space is discrete, Lemma \ref{keylemma} is implied by Theorem 6.2.10 of \citet{Puterman1994}. For completeness, we provide a proof in Appendix \ref{secprooflemma1}  assuming $\mathbb{S}$ belongs to a general vector space. 
In the following, we introduce two variants of MDPs, including HMDPs and POMDPs. These models are illustrated in Figure \ref{fig1}.

\subsection{HMDP}
It can be seen from Figure \ref{fig1} that HMDPs are very similar to MDPs. The difference lies in that in HMDPs, $S_{0,t+1}$ and $R_{0,t}$ depend not only on $(S_{0,t},A_{0,t})$, but $(S_{0,t-1},A_{0,t-1}),\cdots,(S_{0,t-\kappa_0+1},A_{0,t-\kappa_0+1})$ for some integer $\kappa_0>1$ as well. Formally, we have 
\begin{eqnarray}\label{eqHMDP}
\prob (S_{0,t+1}\in \mathcal{S},R_{0,t}\in \mathcal{R}|A_{0,t},\bar{\bm{S}}_{0,t}, \{R_{0,j}\}_{j<t} ) 
=\mathcal{P}( \mathcal{S},\mathcal{R};\{A_{0,j}\}_{t-\kappa_0< j\le t},\{S_{0,j}\}_{t-\kappa_0< j\le t} ),
\end{eqnarray}
for some $\mathcal{P}$, $\kappa_0$ and any $\mathcal{S}\subseteq \mathbb{S}$, $\mathcal{R}\subseteq \mathbb{R}$, $t> \kappa_0$. For any integer $k> 0$, define a new state variable $$S_{0,t}(k)=(S_{0,t}^\top, A_{0,t}, S_{0,t+1}^\top, A_{0,t+1},\cdots,S_{0,t+k-1}^\top)^\top.$$
Let $A_{0,t}(k)=A_{0,t+k-1}$ and $R_{0,t}(k)=R_{0,t+k-1}$ for any $t,k$. 
It follows from \eqref{eqHMDP} that the new process formed by the triplets $(S_{0,t}(\kappa_0),A_{0,t}(\kappa_0),R_{0,t}(\kappa_0))_{t\ge 0}$ satisfies MA. 

For any $k> 0$, let $\hbox{SR}(k)$ denote the set of stationary policies $\pi=\{\pi_t\}_{t\ge 0}$ such that $\pi_t$ depend on $\bar{\bm{S}}_{0,t}$ only through $S_{0,t-k}(k)$. Suppose we are interested in identifying a policy that maximizes the following $k$-step value function
\begin{eqnarray*}
	V^{(k)}(\pi;s)=\sum_{t\ge 0} \gamma^{t} \Mean^{\pi_t} \{R_{0,t}(k)|S_{0,0}(k)=s\},
\end{eqnarray*}
for any $s\in \mathbb{S}(k)$, the state space for $S_{0,t}(k)$. By Lemma \ref{keylemma}, we obtain the following results. 

\begin{lemma}\label{lemma2}
	Assume \eqref{eqHMDP} holds. Then there exists some $\pi^{\tiny{opt}}\in \hbox{SR}(\kappa_0)$ such that 
	$V^{(k)}(\pi^{\tiny{opt}};s)=\sup_{\pi\in \hbox{\scriptsize{HR}}} V^{(k)}(\pi;s)$ for any $s\in \mathbb{S}(k)$ and $k\ge \kappa_0$. 
\end{lemma}

Lemma \ref{lemma2} suggests that in HMDPs, identification of the optimal policy relies on correct specification of the look-back period $\kappa_0$. To determine $\kappa_0$, we can sequentially test whether the triplets $\{(S_{0,t}(k),A_{0,t}(k),R_{0,t}(k))\}_{t\ge 0}$ satisfy MA for $k=1,2,\cdots,$ until the null MA is not rejected. 

\subsection{POMDP}
The POMDP model can be described as follows. At time $t-1$, suppose the environment is in some hidden state $H_{0,t-1}$. The hidden variables $\{H_{0,t}\}_{t\ge 0}$ are unobserved. Suppose the agent chooses an action $A_{0,t-1}$. Similar to MDPs, this will cause the environment to transition to a new state $H_{0,t}$ at time $t$. At the same time, the agent receives an observation $S_{0,t}\in \mathbb{S}$ and a reward $R_{0,t}$ that depend on $H_{0,t}$ and $A_{0,t-1}$. The goal is to estimate an optimal policy based on the observed state-action pairs. 

The observations in POMDPs do not satisfy the Markov property. To better illustrate this, consider the causal diagram for POMDP depicted in Figure \ref{fig1}. The path $S_{0,t-1}\leftarrow H_{0,t-1} \rightarrow H_{0,t} \rightarrow H_{0,t+1}\rightarrow S_{0,t+1}$ connects $S_{0,t-1}$ and $S_{0,t+1}$ without traversing $S_{0,t}$ and $A_{0,t}$. As a result, $S_{0,t+1}$ and $S_{0,t-1}$ are not d-separated \citep[see the definition of d-separation on Page 16,][]{Pearl2000} given $S_{0,t}$ and $A_{0,t}$. Under the faithfulness assumption \citep[see e.g.][]{kalisch2007}, $S_{0,t-1}$ and $S_{0,t+1}$ are mutually dependent conditional on $S_{0,t}$ and $A_{0,t}$. Similarly, we can show $S_{0,t+k}$ and $S_{0,t-1}$ are mutually dependent conditional on $\{(S_{0,j},A_{0,j})\}_{t\le j<t+k}$ for any $k>1$. As a result, the Markov assumption will not hold no matter how many past measurements the state variable includes. This suggests in POMDPs, the optimal policy could be history dependent. 

\section{Testing the Markov assumption}\label{sectest}
\subsection{A CCF-based characterization of MA}
For simplicity, suppose $R_{0,t}$ is a deterministic function of $S_{0,t+1}$, $A_{0,t}$ and $S_{0,t}$. This condition automatically holds if we include $R_{0,t}$ in the set of state variables $S_{0,t+1}$. It is also satisfied in our real dataset (see Section \ref{secrealdata} for details). Under this condition, MA is equivalent to the following,
\begin{eqnarray}\label{MAstar}
\prob (S_{0,t+1}\in \mathcal{S}|A_{0,t},\bar{\bm{S}}_{0,t})
=\mathcal{P}( \mathcal{S};A_{0,t},S_{0,t}),
\end{eqnarray}
for any $\mathcal{S}\subseteq \mathbb{S}$ and $t\ge 0$. Let $\{(S_{1,t},A_{1,t},R_{1,t})\}_{0\le t\le T}$, $\{(S_{2,t},A_{2,t},R_{2,t})\}_{ 0\le t\le T}$, $\cdots$, $\{(S_{n,t},A_{n,t},R_{n,t})\}_{0\le t\le T}$ be i.i.d. copies of $\{(S_{0,t},A_{0,t},R_{0,t})\}_{0\le t\le T}$. Given the observed data, we focus on testing the following hypothesis:

$\hbox{H}_0$: The system is a MDP, i.e, \eqref{MAstar} holds  v.s\\
$\hbox{H}_1$: The system is a HMDP or POMDP. 

In the rest of this section, we present a CCF characterization of $\hbox{H}_0$. For any random vectors $Z_1,Z_2,Z_3$, we use the notation $Z_1\independent Z_2|Z_3$ to indicate that $Z_1$ and $Z_2$ are independent conditional on $Z_3$. To test $H_0$, it suffices to test the following conditional independence assumptions:
\begin{eqnarray}\label{CIA}
S_{0,t}\independent \{(S_{0,j},A_{0,j})\}_{0\le j\le t-2} | S_{0,t-1},A_{0,t-1},\forall t>1.
\end{eqnarray}
For any $t$, let $X_{0,t}=(S_{0,t}^\top,A_{0,t})^\top$ denote the state-action pair. 
For any $\mu \in \mathbb{R}^p$, 
define the following CCF,
\begin{eqnarray}\label{CCF}
\varphi_t(\mu|x)=\Mean \{\exp(i \mu^\top S_{0,t+1})|X_{0,t}=x\}.
\end{eqnarray} 
In the following, we present an equivalent representation for \eqref{CIA} based on \eqref{CCF}. 

%
%


\begin{thm}\label{lemma3}
	\eqref{CIA} is equivalent to the following: for any $t>0$, $q\ge 0$, $\mu \in \mathbb{R}^p$, $\nu \in \mathbb{R}^{p+1}$, we have almost surely,
	\begin{eqnarray}\label{CIA2}
	\varphi_{t+q}(\mu|X_{0,t+q})\Mean [\exp(i\nu^\top X_{0,t-1})|\{X_{0,j}\}_{t\le j\le t+q}]\\\nonumber
	=\Mean [\exp(i\mu^\top S_{0,t+q+1}+i\nu^\top X_{0,t-1})|\{X_{0,j}\}_{t\le j\le t+q}].
	\end{eqnarray}
\end{thm}
Under $H_0$, 
there exists some $\varphi^*$ such that $\varphi_t=\varphi^*$ for any $t$. 
By Theorem \ref{lemma3}, we can show that
\begin{eqnarray*}
	&&\Mean \{\exp(i\mu^\top S_{0,t+q+1})-\varphi^*(\mu|X_{0,t+q}) \}\exp(i\nu^\top X_{0,t-1})\\
	&=&\Mean \exp(i\mu^\top S_{0,t+q+1}+i\nu^\top X_{0,t-1})
	-\Mean \varphi^*(\mu|X_{0,t+q})\exp(i\nu^\top X_{0,t-1})=0,
\end{eqnarray*}
for any $t,q,\mu,\nu$. 
This motivates us to consider the test statistic based on
\begin{eqnarray}\label{est}
\frac{1}{n(T-q-1)}\sum_{j=1}^n \sum_{t=1}^{T-q-1} \{\exp(i\mu^\top S_{j,t+q+1})
-\widehat{\varphi}(\mu|X_{j,t+q}) \}\{\exp(i\nu^\top X_{j,t-1})-\bar{\varphi}(\nu)\},
\end{eqnarray}
where $\widehat{\varphi}$ denotes some nonparametric estimator for $\varphi^*$ and $\bar{\varphi}(\nu)=n^{-1}(T+1)^{-1}\sum_{1\le j\le n,0\le t\le T} \exp(i\nu^\top X_{j,t-1})$.

Modern machine learning (ML) algorithms are well-suited to estimating $\varphi^*$ in high-dimensional cases. 
However, naively plugging ML estimators for $\widehat{\varphi}$ will cause a heavy bias in \eqref{est}. Because of that, the resulting estimating equation does not have a tractable limiting distribution. Kernel smoothers \citep{Hardle1990} or local polynomial regression can be used to reduce the estimation bias by properly choosing the bandwidth parameter. However, as commented in Section \ref{secrelatedwork}, these methods suffer from the curse of dimensionality and will perform poorly in 
cases as we concatenate data over multiple decision points. 

In the next section, we address these concerns by presenting a doubly-robust estimating equation to alleviate the estimation bias. When observations are time independent, our method shares similar spirits with 
the double machine learning method proposed by \citet{Cherno2018} for statistical inference of the average treatment effects in causal inference.  


\subsection{Forward-Backward Learning}\label{secFBML}
To introduce our method, we define another CCF
\begin{eqnarray}\label{anotherCCF}
\psi_t(\nu|x)=\Mean \{\exp(i \nu^\top X_{0,t-1})|X_{0,t}=x\}.
\end{eqnarray} 
We need the following two conditions. 

\noindent (C1) 
Actions are generated by a fixed behavior policy. \\
\noindent (C2) Suppose the process $\{S_{0,t}\}_{t\ge 0}$ is strictly stationary.  

Condition (C1) requires the agent to select actions based on information contained in the current state variable only. 
Under $H_0$, the process $\{S_{0,t}\}_{t\ge 0}$ forms a time-invariant Markov chain. When its initial distribution equals its stationary distribution, (C2) is automatically satisfied. This together with (C1) implies $\{X_{0,t}\}_{t\ge 0}$ is strictly stationary as well. As a result, we have $\psi_t=\psi^*$ for some $\psi^*$ and any $t>0$. 

\begin{thm}\label{lemma4}
	Suppose $H_0$, (C1) and (C2) hold. Then for any $t>0$, $q\ge 0$, $\mu\in \mathbb{R}^p$, $\nu \in \mathbb{R}^{p+1}$, we have
	\begin{eqnarray*}
		\Mean \Gamma_0(q,\mu,\nu)\equiv \Mean \{\exp(i\mu^\top S_{0,t+q+1})-\varphi^*(\mu|X_{0,t+q}) \}
	 \{\exp(i\nu^\top X_{0,t-1})-\psi^*(\nu|X_{0,t}) \}=0.
	\end{eqnarray*}
	Moreover, the above equation is doubly-robust. That is, for any CCFs $\varphi$ and $\psi$, the following holds as long as either $\varphi=\varphi^*$ or $\psi=\psi^*$,
	\begin{eqnarray}\label{equation}
	\Mean \{\exp(i\mu^\top S_{0,t+q+1})-\varphi(\mu|X_{0,t+q}) \} \{\exp(i\nu^\top X_{0,t-1})-\psi(\nu|X_{0,t}) \}=0.
	\end{eqnarray}
\end{thm}
\textbf{Proof: }\textit{When $\varphi=\varphi^*$, 
	we have
	\begin{eqnarray*}
		\Mean [\exp(i\mu^\top S_{0,t+q+1})-\varphi^*(\mu|X_{0,t+q})|\{X_{0,j}\}_{ j\le t+q}]=0,
	\end{eqnarray*}
	under MA. Assertion \eqref{equation} thus follows. Under (C1), we have $X_{0,t-1}\independent \{X_{0,j}\}_{j>t}|X_{0,t}$ for any $t>1$. When $\psi=\psi^*$, we can similarly show that
	\begin{eqnarray*}
		\Mean [\exp(i\nu^\top X_{0,t-1})-\psi^*(\nu|X_{0,t})|\{X_{0,j}\}_{j>t} ]=0.
	\end{eqnarray*}
	The doubly-robustness property thus follows. }

The propose algorithm estimates both $\varphi^*$ and $\psi^*$ using ML methods without specifying their parametric forms. 
Let $\widehat{\varphi}$ and $\widehat{\psi}$ denote the corresponding estimators. Note that computing $\varphi^*$ is essentially estimating the characteristic function of $S_{0,t}$ given $S_{0,t-1}$. This corresponds to a forward prediction task. Similarly, estimating $\psi^*$ is a backward prediction task. Thus, we refer to $\widehat{\varphi}$ and $\widehat{\psi}$ as \textbf{forward} and \textbf{backward} \textbf{learners}, respectively. Our proposed method is referred to as the \textbf{forward-backward learning} algorithm. It is worth mentioning that although we focus on the problem of testing MA in this paper, the proposed method can be applied to more general estimation and inference problems with time-dependent observations.

Consider the following estimating equation,
\begin{eqnarray}\label{est2}
\frac{1}{n(T-q-1)}\sum_{j=1}^n \sum_{t=1}^{T-q-1} \{\exp(i\mu^\top S_{j,t+q+1})
-\widehat{\varphi}(\mu|X_{j,t+q}) \} \{\exp(i\nu^\top X_{j,t-1})-\widehat{\psi}(\nu|X_{j,t})\}.
\end{eqnarray}
Unlike \eqref{est}, the above estimating equation is doubly robust. This helps alleviate the impact of the biases in $\widehat{\varphi}$ and $\widehat{\psi}$. 

Our test statistic is constructed based on a slightly modified version of \eqref{est2} with cross-fitting. The use of cross-fitting allows us to establish the limiting distribution of the estimating equation under minimal conditions. 

Suppose we have at least two trajectories, i.e, $n\ge 2$. 
We begin by randomly dividing $\{1,\cdots,n\}$ into $\mathbb{L}$ subsets $\mathcal{I}^{(1)},\cdots,\mathcal{I}^{(\mathbb{L})}$ of equal size. Denote by $\mathcal{I}^{(-\ell)}=\{1,\cdots,n\}-\mathcal{I}^{(\ell)}$ for $\ell=1,\cdots,\mathbb{L}$. Let $\widehat{\varphi}^{(-\ell)}$ and $\widehat{\psi}^{(-\ell)}$ denote the forward and backward learners based on the data in $\mathcal{I}^{(-\ell)}$. For any $\mu,\nu,q$, define
\begin{eqnarray*}
	\widehat{\Gamma}(q,\mu,\nu)=\frac{n^{-1}}{T-q-1} \sum_{\ell=1}^{\mathbb{L}} \sum_{j\in \mathcal{I}^{(\ell)}}\sum_{t=1}^{T-q-1}\{\exp(i\mu^\top S_{j,t+q+1})
	-\widehat{\varphi}^{(-\ell)}(\mu|X_{j,t+q}) \} \{\exp(i\nu^\top X_{j,t-1})-\widehat{\psi}^{(-\ell)}(\nu|X_{j,t})\}.
\end{eqnarray*}
Notice that $\widehat{\Gamma}$ is a complex-valued function. We use $\widehat{\Gamma}_R$ and $\widehat{\Gamma}_I$ to denote its real and imaginary part. 

\begin{algorithm}
	\caption{Forward-Backward Learning}\label{alg1}
	\begin{algorithmic}
		\STATE \textbf{Input:} $B$, $Q$, $\mathbb{L}$, $\alpha$ and the observed data.
		\STATE \textbf{Step 1: }Randomly generate i.i.d. pairs $\{(\mu_b,\nu_b)\}_{1\le b\le B}$ from $N(0, I)$; Randomly divide $\{1,\cdots,n\}$ into $\bigcup_{\ell}\mathcal{I}^{(\ell)}$ for $\ell=1,\cdots,\mathbb{L}$, set $\mathcal{I}^{(-\ell)}=\{1,\cdots,n\}-\mathcal{I}^{(\ell)}$. 
		\STATE \textbf{Step 2: }Compute the forward and backward learners $\widehat{\varphi}^{(-\ell)}(q,\mu_b,\cdot)$ and $\widehat{\psi}^{(-\ell)}(q,\nu_b,\cdot)$ for $q=0,\cdots,Q$, $b=1,\cdots,B$ based on modern ML methods.
		\STATE \textbf{Step 3: }Compute $\widehat{\Gamma}(q,\mu_b,\nu_b)$ for $q=0,\cdots,Q$, $b=1,\cdots,B$; Compute $\widehat{S}$ according to \eqref{TS}.
		\STATE \textbf{Step 4: }For $q=0,\cdots,Q$, compute an estimated covariance matrix $\widehat{\Sigma}^{(q)}$ according to \eqref{sigmaq} (see Appendix \ref{secsigma} for details).
		\STATE \textbf{Step 5: }Use Monte Carlo to simulate the upper $\alpha/2$-th critical value of $\max_{q\in \{0,\dots,Q\}} \|\{\widehat{\bm{\Sigma}}^{(q)}\}^{1/2} \mathbb{Z}_q\|_{\infty}$ where $\mathbb{Z}_2,\cdots,\mathbb{Z}_Q$ are i.i.d. $2B$-dimensional random vectors with identity covariance matrix. Denote this critical value by $\widehat{c}_{\alpha}$. 
		\STATE \textbf{Reject} $\hbox{H}_0$ if $\widehat{S}$ is greater than $\widehat{c}_{\alpha}$.
	\end{algorithmic}	
\end{algorithm}	

To implement our test, we randomly sample i.i.d. pairs $\{(\mu_b,\nu_b)\}_{1\le b\le B}$ according to a multivariate normal distribution with zero mean and identity covariance matrix, where $B$ is allowed to diverge with the number of observations. Let $Q$ be some large integer that is allowed to be proportion to $T$ (see the condition in Theorem \ref{thm1} below for details). We calculate $\widehat{\Gamma}_R(q,\mu_b,\nu_b)$ and $\widehat{\Gamma}_I(q,\mu_b,\nu_b)$ for $b=1,\cdots,B$, $q=0,\cdots,Q$. Under $H_0$, 
$\widehat{\Gamma}_R(q,\mu_b,\nu_b)$ and $\widehat{\Gamma}_I(q,\mu_b,\nu_b)$ are close to zero. Thus, we 
reject $H_0$ when one of these quantities has large absolute value. Our test statistic is given by
\begin{eqnarray}\label{TS}
\widehat{S}=\max_{b\in \{1,\cdots,B\}}\max_{q\in \{0,\cdots,Q\}} \sqrt{n(T-q-1)}
\max(|\widehat{\Gamma}_R(q,\mu_b,\nu_b)|,|\widehat{\Gamma}_I(q,\mu_b,\nu_b)|).
\end{eqnarray}
Under $H_0$, each $\widehat{\Gamma}_R(q,\mu_b,\nu_b)$ (or $\widehat{\Gamma}_I(q,\mu_b,\nu_b)$) is asymptotically normal. As a result, $\widehat{S}$ converges in distribution to a maximum of some Gaussian random variables. For a given significance level $\alpha>0$, we reject $H_0$ when $\widehat{S}>\widehat{c}_{\alpha}$ for some threshold $\widehat{c}_{\alpha}$ computed by wild bootstrap \citep{Wu1986}. 
We detail our procedure in Algorithm \ref{alg1}.

Step 2 of our algorithm requires to estimate $\widehat{\varphi}^{(-\ell)}(\mu_b|\cdot)$ and $\widehat{\psi}^{(-\ell)}(\nu_b|\cdot)$ for $b=1,\cdots,B$. The integer $B$ shall be large enough to guarantee that our test has good power properties. Our method allows $B$ to grow at an arbitrary polynomial order of $n\times T$ (see the condition in Theorem \ref{thm1} below for details). Separately applying ML algorithms $B$ times to compute these leaners is computationally intensive. 
In Section \ref{secimp}, we use the random forests \citep{breiman2001random} algorithm as an example to illustrate how these leaners can be simultaneously calculated. Other ML algorithms could also be used. 

\subsection{Bidirectional asymptotics}\label{secBA}
In this section, we prove the validity of our test under a bidirectional-asymptotic framework where either $n$ or $T$ grows to infinity. We begin by introducing some conditions. 

(C3) Under $H_0$, suppose the Markov chain $\{X_{0,t}\}_{t\ge 0}$ is geometrically ergodic when $T\to \infty$. \\ 
(C4) 
Suppose there exists some $c_0>1/2$ such that
\begin{eqnarray*}
	\max_{1\le b\le B}  \int_{x}|\widehat{\varphi}^{(-\ell)}(\mu_b|x)-\varphi^*(\mu_b|x)|^2\mathbb{F}(dx)=O_p((nT)^{-c_0}),\\
	\max_{1\le b\le B} \int_{x} |\widehat{\psi}^{(-\ell)}(\nu_b|x)-\psi^*(\nu_b|x)|^2\mathbb{F}(dx)=O_p((nT)^{-c_0}),
\end{eqnarray*}
where $\mathbb{F}$ denotes the distribution function of $X_{0,0}$. In addition, suppose $\widehat{\varphi}^{(-\ell)}$ and $\widehat{\psi}^{(-\ell)}$ are bounded functions.

Condition (C3) enables us to establish the limiting distribution of our test under the setting where $T\to \infty$. Notice that this condition is not needed when $T$ is bounded. The geometric ergodicity assumption \citep[see e.g.][for  definition]{Luke1994} is weaker than the uniform ergodicity condition imposed in the existing reinforcement learning literature \citep[see e.g.][]{bhandari2018,zou2019}. There exist Markov chains that are not uniformly ergodic but may still be geometrically ergodic \citep{Mengersen1996}.

The first part of Condition (C4) requires the prediction errors of estimated CCFs to satisfy certain uniform convergence rates. This is the key condition to ensure valid control of the type-I error rate of our test. In practice, the capacity of modern ML algorithms and their success in prediction tasks even in
high-dimensional samples make this a reasonable assumption. In theory, the uniform convergence rates in (C4) can be derived for popular ML methods such as random forests \citep{Biau2012} and deep neural networks \citep{schmidt2020}. The boundedness assumption in (C4) is reasonable since $\varphi^*$ and $\psi^*$ are bounded by $1$.  

\begin{thm}\label{thm1}
	Assume (C1)-(C4) hold. Suppose $\log B=O((nT)^{c^*})$ for any finite $c^*>0$ and $Q\le \max(\rho_0 T,T-2)$ for some constant $\rho_0<1$. In addition, suppose there exists some $\epsilon_0>0$ such that the real and imaginary part of $\Gamma_0(q,\mu,\nu)$ 
	have variances greater than $\epsilon_0$ for any $\mu,\nu$ and $q\in \{0,\cdots,Q\}$. Then we have as either $n\to \infty$ or $T\to \infty$, $\prob(\widehat{S}>\widehat{c}_{\alpha})=\alpha+o(1)$.
\end{thm}

Theorem \ref{thm1} implies the type-I error rate of our test is well-controlled. Our proof relies on the high-dimensional martingale central limit theorem that is recently developed by \citet{belloni2018}. This enables us to show the asymptotic equivalence between the distribution of $\widehat{S}$ and that of the bootstrap samples given the data, under settings where $B$ diverges with $n$ and $T$. 
It is worthwhile to mention that the stationarity condition in (C2) is imposed to simplify the presentation. Our test remains valid when (C2) is violated. To save space, we move the related discussions to Appendix \ref{secwithoutstationary}. 




\section{Model selection}\label{modelselect}
\begin{algorithm}
	\caption{RL Model Selection}\label{alg2}
	\begin{algorithmic}
		\STATE \textbf{Input:} $B$, $Q$, $\mathbb{L}$, $\alpha$ and the observed data.
		\FOR{$k=1,2,\cdots,K$} 
		\STATE \textbf{Apply} algorithm \ref{alg1} with $B$, $Q$, $\mathbb{L}$, $\alpha$ specified above to the data $\{(S_{j,t}(k),A_{j,t}(k))\}_{1\le j\le n,0\le t\le T-k+1}$.
		\IF{$\hbox{H}_0$ is not rejected}
		\STATE \textbf{Conclude} the system is a $k$-th order MDP; \textbf{Break}.
		\ENDIF
		\ENDFOR
		\STATE \textbf{Conclude} the system is a POMDP.
	\end{algorithmic}	
\end{algorithm}	
Based on our test, we can choose which RL model to use to model the system dynamics. For any $j,k,t$, let
$$S_{j,t}(k)=(S_{j,t}^\top, A_{j,t}, S_{j,t+1}^\top, A_{j,t+1},\cdots,S_{j,t+k}^\top)^\top,$$
and $A_{j,t}(k)=A_{j,t+k}$. Given a large integer $K$, our procedure sequentially test the null hypothesis MA based on the concatenated data $\{(S_{j,t}(k),A_{j,t}(k))\}_{1\le j\le n,0\le t\le T-k}$ for $k=0,1,\cdots,K$. Once the null is not rejected, we can conclude the system is a $k$-th order MDP and terminate our procedure. Otherwise, we conclude the system is a POMDP. We summarize our method in Algorithm \ref{alg2}.


\section{Numerical examples}\label{secnumerical}
This section is organized as follows. We discuss some implementation details in Section \ref{secimp}. In Section \ref{sec:ohio}, we apply our test to mobile health applications. We use both synthetic and real datasets to demonstrate the usefulness of our test in detecting HMDPs. In Section \ref{sec:tiger}, we apply our test to a POMDP problem to illustrate its consistency. 

\subsection{Implementation details}\label{secimp}
We first describe the algorithm we use to simultaneously compute $\{\widehat{\varphi}^{(-\ell)}(\mu_b|\cdot)\}_{1\le b\le B}$. 
The algorithm for computing backward learners can be similarly derived. 
Our method is motivated by the quantile regression forest algorithm \citep{mein2006}. We detail our procedure below. 
\%vspace{-0.4cm}
\begin{enumerate}
	\item Apply the random forests algorithm with the response-predictor pairs $\{(S_{j,t},X_{j,t-1})\}_{j\in \mathcal{I}^{(-\ell)},1\le t\le T}$  to grow $M$ trees $T(\theta_m)$ for $m = 1, \dots, M$. Here $\theta_m$ denotes the parameters associated with the $m$-th tree. Denote by
	$l(x,\theta_m)$ the leaf space of the $m$-th tree that predictor $x$ fails into.
	\%vspace{-0.2cm}
	\item For any $m\in \{1,\cdots,T\}$, $(j,t)\in \mathcal{I}^{(-\ell)}$ and $x$, compute the weight  $w_{j,t}^{(-\ell)}(x,\theta_m)$ as $$\frac{\mathbb{I}\{X_{j,t} \in l(x,\theta_m )\}}{ \#\{(l_1,l_2): l_1\in \mathcal{I}^{(-\ell)},X_{l_1,l_2} \in l(x,\theta_m)\} }.$$
	Average over all trees to calculate the weight of each training data as $w_{j,t}^{(-\ell)}(x) = \sum_{m = 1}^M w_{j,t}^{(-\ell)}(x,\theta_m) / M$. 
	\item For any $x$ and $b \in \{1, \dots, B\}$, compute the forward learner $\widehat{\varphi}^{(-\ell)}(\mu_b|x)$
	as  the weighted average $\sum_{j\in \mathcal{I}^{(-\ell)},1\le t\le T} w_{j,t}^{(-\ell)}(x) \exp(i \mu_b^\top S_{j,t})$. 
\end{enumerate}

To implement this algorithm, the number of trees $M$ is set to 100 and other tuning parameters are selected via 5-fold cross-validation. 
To construct our test, the hyperparameters $B$, $Q$ and $\mathbb{L}$ are fixed as $100$, $8$ and $3$ respectively. 
All state variables are normalized to have unit sampling variance before running the test. Normalization will not affect the Type I error rate of our test but helps improve its power. Our experiments are run on an c5d.24xlarge instance on the AWS EC2 platform, with 96 cores and 192GB RAM.


\subsection{Applications in HMDP problems}\label{sec:ohio}
\subsubsection{THE OHIOT1DM Dataset}\label{secrealdata}
There has been increasing interest in applying RL algorithms to mobile health (mHealth) applications. 
In this section, we use the OhioT1DM dataset \cite{marling2018ohiot1dm} as an example to illustrate the usefulness of test in mHealth applications. The data contains continuous measurements for six patents with type 1 diabetes over eight weeks. In order to apply RL algorithms, it is crucial to determine how many lagged variables we should include to construct the state vector. 

In our experiment, we divide each day of follow-up into one hour intervals and a treatment decision is made every hour. We consider three important time-varying variables to construct $S_{0,t}$, including the average blood glucose levels $\hbox{G}_{0,t}$ during the one hour interval $(t-1,t]$, the carbohydrate estimate for the meal $\hbox{C}_{0,t}$ during $(t-1,t]$ and $\hbox{Ex}_{0,t}$ which measures exercise intensity during $(t-1,t]$. At time $t$, we define $A_{0,t}$ by discretizing the amount of insulin $\hbox{In}_{0,t}$ injected and define $R_{0,t}$ according to the Index of Glycemic Control \citep{rodbard2009interpretation} that is a deterministic function $\hbox{G}_{0,t+1}$. To save space, we present detailed definitions of $A_{0,t}$ and $R_{0,t}$ in Appendix \ref{secAtRt}.


\subsubsection{synthetic data}\label{sec:ohio_simu}
We first simulate patients with type I diabetes to mimic the OhioT1DM dataset. According to our findings in Section \ref{sec:real_ohio}, we model this sequential decision problem by a fourth order MDP. 
Specifically, we consider the following model for $\hbox{G}_{0,t}$:
\begin{equation*}
\hbox{G}_{0,t} = \alpha + \sum_{i = 1}^{4}(\boldsymbol{\beta}_i^T S_{0,t - i} + c_i A_{0,t-i}) + E_{0,t}, 
\end{equation*}
where $\alpha$, $\{\boldsymbol{\beta}_i\}_{i = 1}^4$ and $ \{c_i\}_{i = 1}^4$ are computed by least-square estimation based on the OhioT1DM dataset. The error term $E_{0,t}$ is set to follow $N(0, 9)$. 

At each time point, a patient randomly choose to consume food with probability $p_1$ and take physical activity with probability $p_2$, where the amounts and intensities are independently generated from normal distributions. The initial values of $\hbox{G}_{0,t}$ are also randomly sampled from a normal distribution.
Actions are independently generated from a multinoulli distribution. 
Parameters 
$p_1,p_2$ as well as other parameters in the above distributions are all estimated from the data. 

For each simulation, we generate $N=10,15$ or $20$ trajectories according to the above model. For each trajectory, we generate measurements with $T=1344$ time points (8 weeks) after an initial burn-in period of $10$ time points. 
For $k\in \{1, \dots, 10\}$, 
we use our test to determine whether the system is a $k$-th order MDP. Under our generative model, we have $H_0$ holds when $k\ge 4$ and $H_1$ holds otherwise.  

\%vspace{-0.2cm}
\begin{figure}[h!]
	 \centering
	\includegraphics[width=0.30\linewidth]{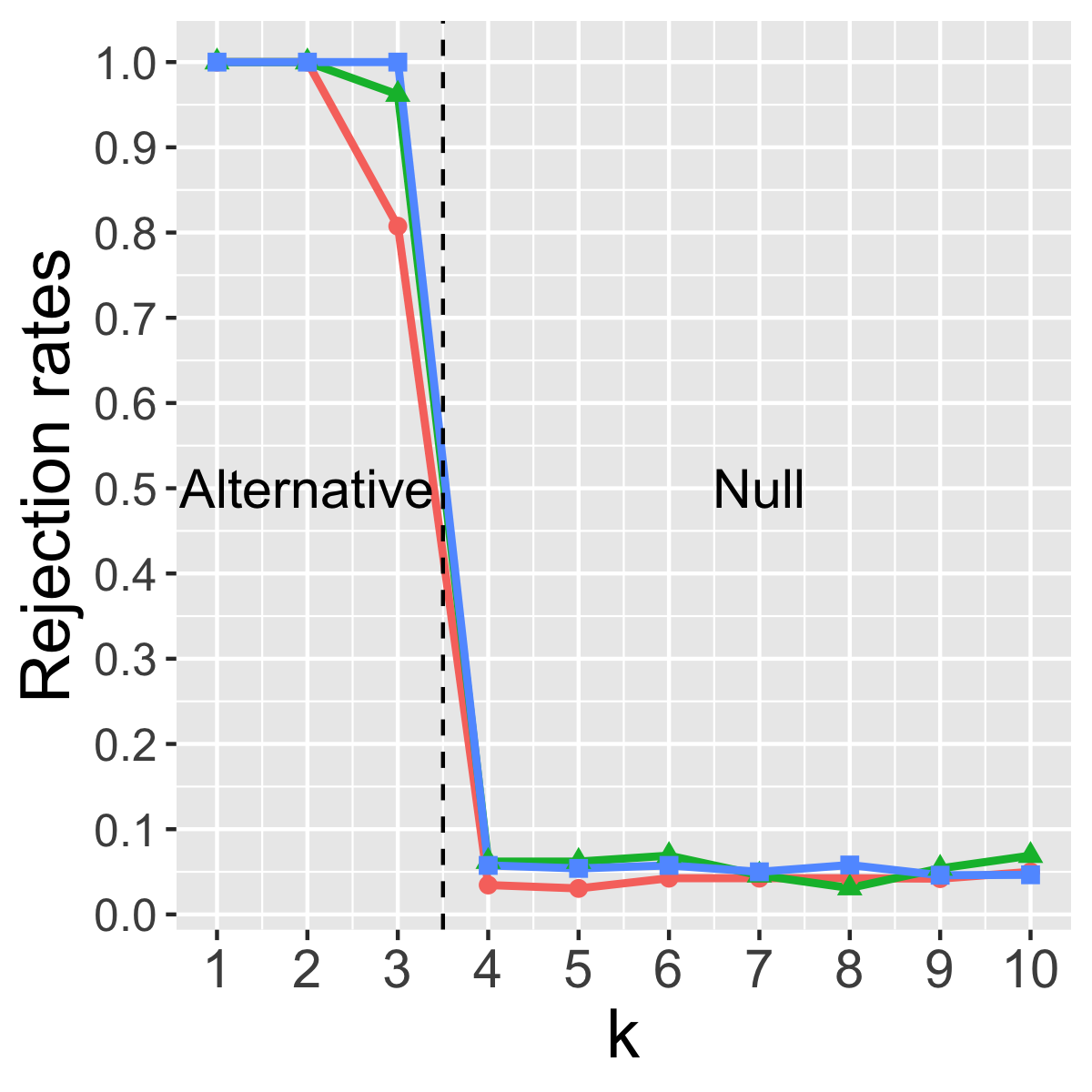}
  \centering
	\includegraphics[width=0.36\linewidth]{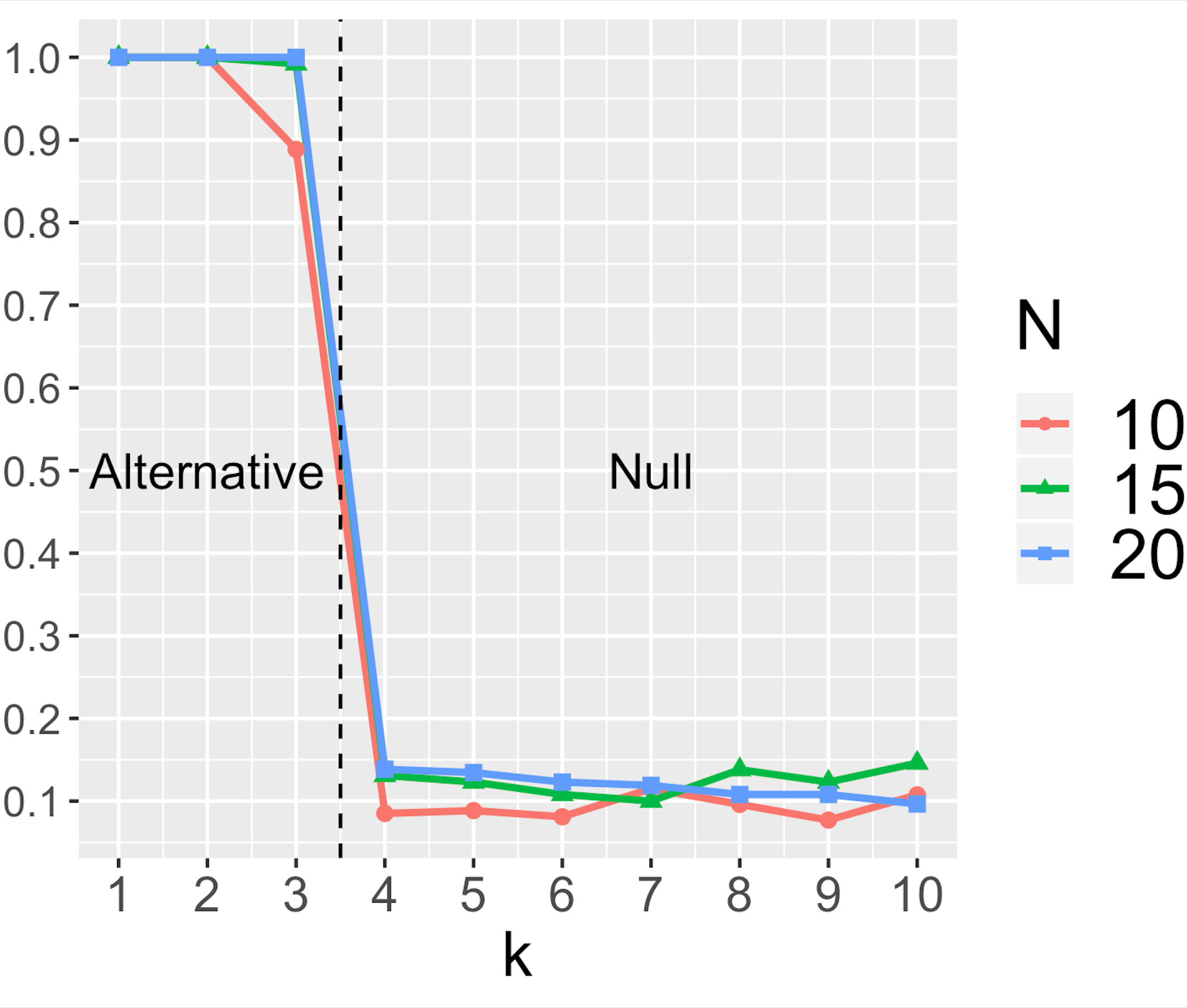}
	\caption{Empirical rejection rates aggregated over 500 simulations with different combinations of $\alpha$, $N$ and $k$. $\alpha=(0.05,0.1)$ from left plot to right plot.
	}
	\label{fig:simu_ohio_test}
	%
	\centering
	\includegraphics[width=0.45\linewidth]{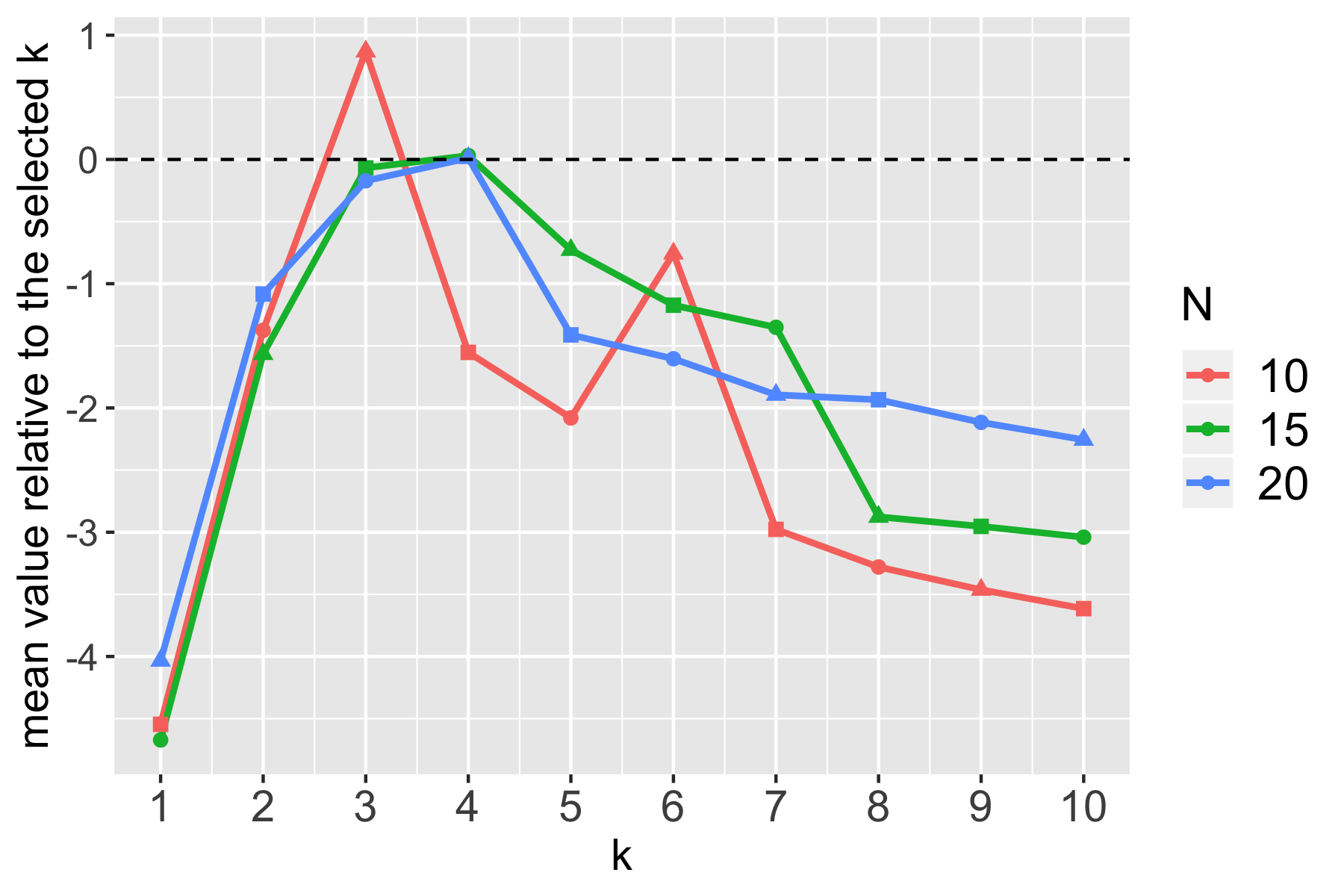}
	\caption{ Value differences with different combinations of $k$ and $N$. }
	\label{fig:simu_ohio_value}
\end{figure}
Empirical rejection rates of our test with different combinations of $k$, $N$ and the significance level $\alpha$ are reported in Figure \ref{fig:simu_ohio_test}. Results are aggregated over 500 simulations. It can be seen that the Type I error rate of our test is close to the nominal level in almost all cases. In addition, its power increases with $N$, demonstrating the consistency of our test.

\begin{table*}[h]\centering 
	\small
	\caption{Policy evaluation results for the OhioT1DM dataset. }
	\vspace{0.1 in}
	\label{tab: real_Ohio}
	\begin{tabular}{|l|l|l|l|l|l|l|l|l|l|l|}
		\hline
		k & 1 & 2 & 3 & 4 & 5 & 6 & 7 & 8 & 9 & 10 \\ \hline
		Estimated value $\bar{V}_k$ & -90.82 & -57.53 & -63.77 & \textbf{-52.57} & -56.23 & -60.05 & -63.70 & -54.85 & -65.08 & -59.59 \\ \hline
	\end{tabular}
\end{table*}

To further illustrate the usefulness of our test, 
we apply Algorithm \ref{alg2} with $\alpha=0.01$, $K=10$ for model selection and evaluate the policy learned based on the selected model. Specifically, let $\widehat{\kappa}_0^{(l)}$ denote the order of MDP estimated by Algorithm \ref{alg2} in the $l$-th simulation. For each $k\in \{1,\cdots,10\}$, we apply the fitted-Q iteration algorithm \citep[][see Section \ref{supp:num_simu} for details]{ernst2005tree} to the data $\{S_{j,t}(k),A_{j,t}(k),R_{j,t}(k)\}_{1\le j\le N, 0\le t\le T-k+1}$ generated in the $l$-th simulation to learn an optimal policy $\widehat{\pi}^{(l)}(k)$ and then simulate 100 trajectories following $\widehat{\pi}^{(l)}(k)$ to compute the average discounted reward $V^{(l)}(k)$ (see Appendix \ref{supp:num_simu} for details). Finally, for each $k=1,\cdots,10$, we compute the value difference
\begin{eqnarray*}
	\hbox{VD}(k)=\frac{1}{500}\sum_{l=1}^{500} \{V^{(l)}(k)-V^{(l)}(\widehat{\kappa}_0^{(l)})\},
\end{eqnarray*}
to compare the policy learned based on our selected model with those by assuming the system is a $k$-th order MDP. 
We report these value differences with different choices of $N$ in Figure \ref{fig:simu_ohio_value}. It can be seen that $\hbox{VD}(k)$ is smaller than or close to zero in almost all cases. When $k=4$, the value differences are very close to zero for large $N$. This suggests that our method is useful in identifying the optimal policy in HMDPs. 
%

\subsubsection{real data analysis}\label{sec:real_ohio}
The lengths of trajectories in the OhioT1DM dataset range from 
1119 to 1288. To implement our test, we set $T=1100$ and apply Algorithm \ref{alg1} to test whether the system is a $k$-th order MDP. The corresponding p-values are reported in Table \ref{tab: real_Ohio}. To apply Algorithm \ref{alg2} for model selection, we set $\alpha=0.01$. Our algorithm stops after the fourth iteration. The first four p-values are 0, 0, 0.001 and 0.068, respectively. 
Thus, we conclude the system is a $4$-th order MDP. 

Next, we use cross-validation to evaluate our selected model. Specifically, 
we split the six trajectories into training and testing sets, 
with each containing three trajectories. This yields a total of $L={6\choose 3}=20$ combinations. 
Then for each combination and $k\in \{1,\cdots,10\}$, we apply FQI to learn an optimal policy based on the training dataset by assuming the system is a $k$-th order MDP and apply the Fitted Q evaluation algorithm \cite{le2019batch} on the testing dataset to evaluate its value (see Appendix \ref{supp:num_real} for details). Finally, we aggregated these values over different combinations and report them in Table \ref{tab: real_Ohio}. 
It can be seen that the policy learned based on our selected model achieves the largest value.

\subsection{Applications in POMDP problems}\label{sec:tiger}
We apply our test to the Tiger problem \citep{cassandra1994acting}. 
The model is defined as follows: 
at the initial time point, a tiger is randomly placed behind either the left or the right door with equal probability. 
At each time point, the agent can select from one of the following three actions: (i) open the left door; (ii) open the right; (iii) listen for tiger noises. But listening is not entirely accurate. If the agent chooses to listen, it will receive an observation $S_{0,t}$ that corresponds to the estimated location of the tiger. Let $H_{0,t}$ denote the observed correct location of the tiger, we have $\prob(H_{0,t}=S_{0,t})=0.7$ and $\prob(H_{0,t}\neq S_{0,t})=0.3$. 
If the agent chooses to open one of two doors, it receives a penalty of -100 if the tiger is behind that door or a reward $R_{0,t}$ of +10 otherwise. The game is then terminated. 

\begin{figure}[!h]
 \centering
	\includegraphics[width=0.33\linewidth]{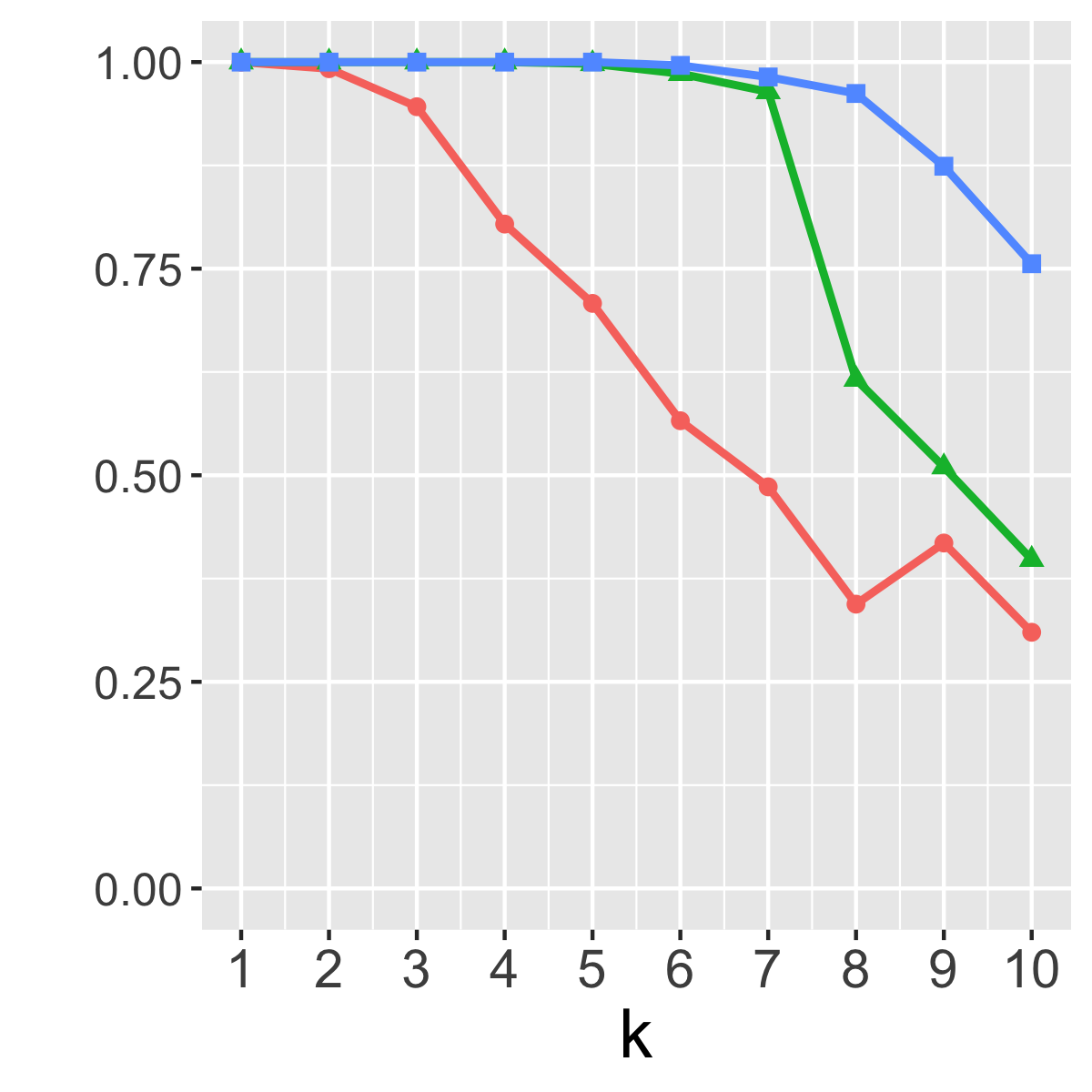}
	\includegraphics[width=0.42\linewidth]{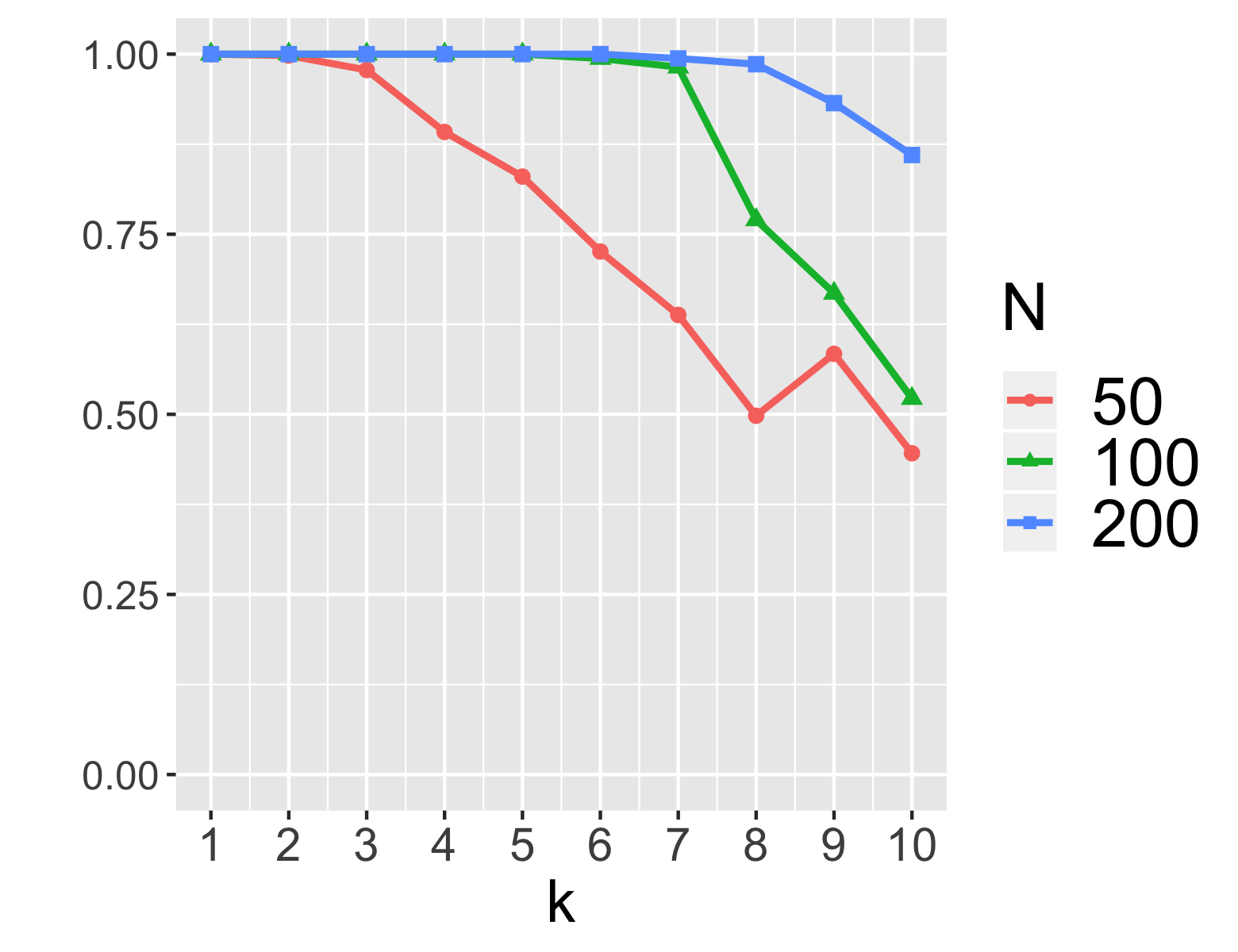}
	\includegraphics[width=0.33\linewidth]{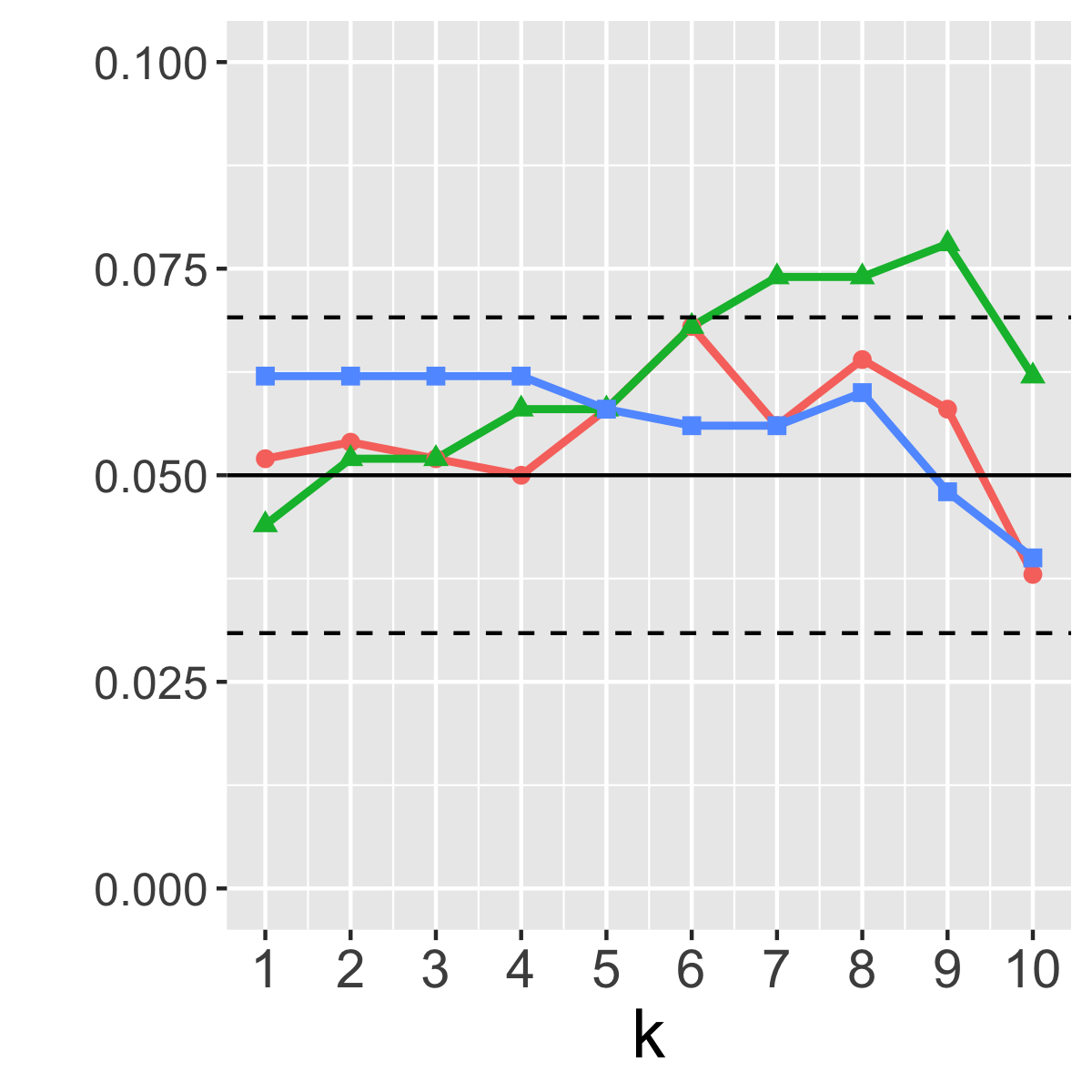}
	\includegraphics[width=0.42\linewidth]{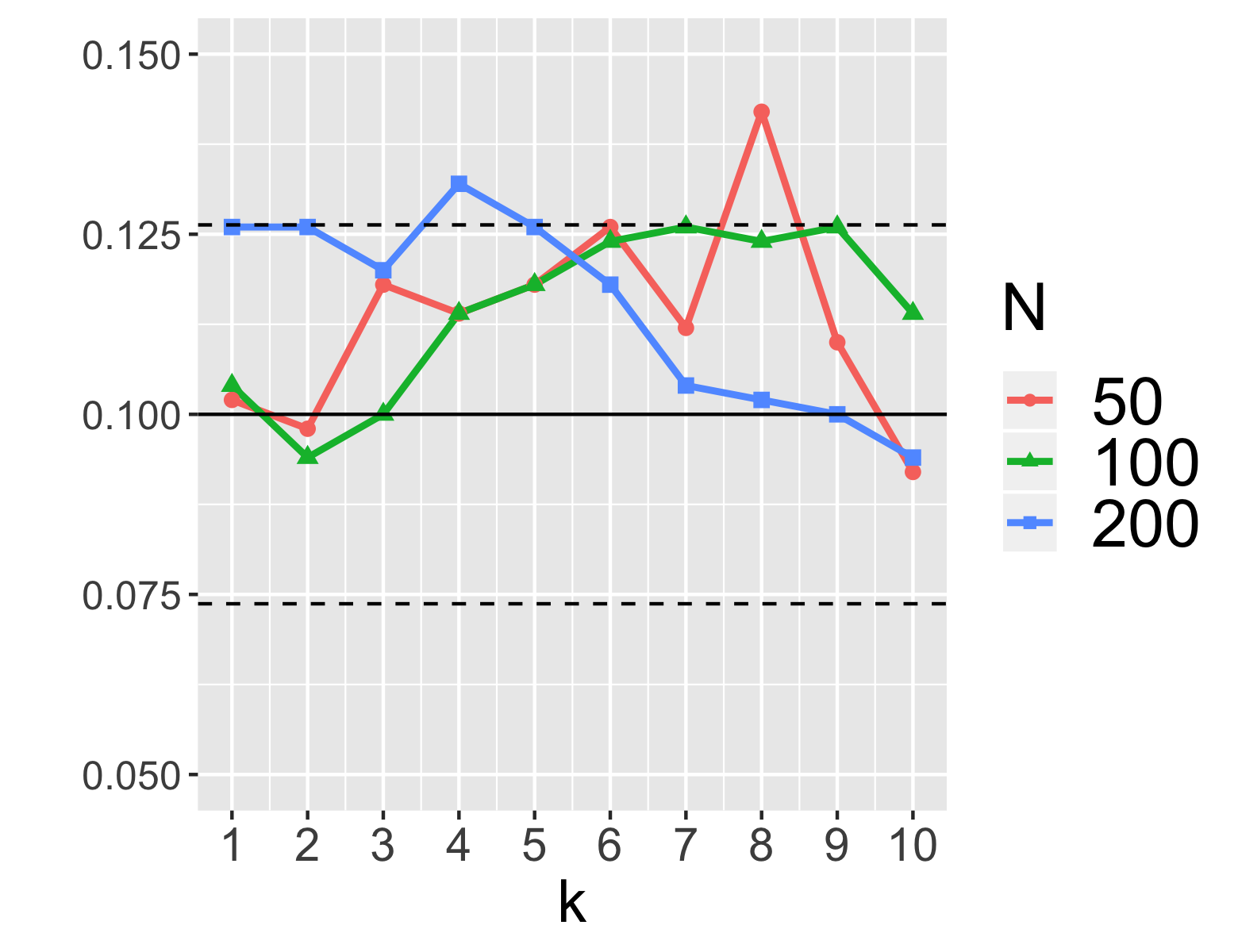}
	\caption{Empirical rejection rates aggregated over 500 simulations with different combinations of $\alpha$, $K$ and $N$. $\alpha=(0.05,0.1)$ from left plots to right plots. $H_1$ holds in top plots. $H_0$ holds in bottom plots. Dashed lines correspond to $y=\alpha\pm 1.96 \hbox{MCE}$ where MCE denotes the Monte Carlo error $\sqrt{\alpha(1-\alpha)/500}$. 
	}
	\label{fig:simu_tiger_1}
\end{figure}

We set $T$ to $20$. To generate the data, the behaviour policy is set to listening at time points $t=0,1,2,\cdots,T-1$ and randomly choosing a door to open with equal probability at time $T$. 
For each simulation, we generate a total of $N$ trajectories and then
apply Algorithm \ref{fig1} to
the data $\{(S_{j,t}(k),A_{j,t}(k))\}_{1\le j\le N,0\le t\le T-k+1}$
for $k = 1, \dots, 10$. 
The empirical rejection rates with $N=50,100$ and $200$ and the significance level $\alpha=0.05$ and $0.1$ are reported in the top plots of Figure \ref{fig:simu_tiger_1}. It can be seen that our test has nonnegligible powers for detecting POMDPs. Take $\alpha=0.1$ as an example. The rejection rate is well above $50\%$ in almost all cases. Moreover, the power of our test increases as either $N$ increases or $k$ decreases, as expected.

To evaluate the validity our test in this setting, we define a new state vector $S_{0,t}^*=(S_{0,t},H_{0,t})^\top$ and repeat the above experiment with this new state. Since the hidden variable is included in the state vector, the Markov property is satisfied. 
The empirical rejection rates with different combinations of $N$, $\alpha$ and $k$ are reported in the bottom plots of Figure \ref{fig:simu_tiger_1}. It can be seen that the Type I error rates are well-controlled in almost all cases. 
\bibliography{TestMA}
\bibliographystyle{icml2020}

\appendix
\section{Additional details regarding our test}
\subsection{The covariance estimator $\widehat{\Sigma}^{(q)}$}\label{secsigma}
For any $\ell =1,\cdots,\mathbb{L}$, $j\in \mathcal{I}^{(\mathbb{\ell})}$ and $0< t<T-q$, define vectors $\lambda_{R,q,j,t},\lambda_{I,q,j,t}\in \mathbb{R}^{\mathbb{B}}$ such that the $b$-th element of $\lambda_{R,q,j,t},\lambda_{I,q,j,t}$ correspond to the real and imaginary part of 
\begin{eqnarray*}
	\{\exp(i\mu^\top S_{j,t+q+1})
	-\widehat{\varphi}^{(-\ell)}(\mu|X_{j,t+q}) \} \{\exp(i\nu^\top X_{j,t-1})-\widehat{\psi}^{(-\ell)}(\nu|X_{j,t})\},
\end{eqnarray*}
respectively. The matrix $\widehat{\Sigma}^{(q)}$ is defined by
\begin{eqnarray}\label{sigmaq}
\sum_{\ell=1}^{\mathbb{L}} \sum_{j\in \mathcal{I}^{(\ell)}}\sum_{t=1}^{T-q-1} \frac{(\lambda_{R,q,j,t}^\top,\lambda_{I,q,j,t}^\top)^\top  (\lambda_{R,q,j,t}^\top,\lambda_{I,q,j,t}^\top)}{n(T-q-1)}.
\end{eqnarray}
\subsection{Validity of our test without the stationary assumption}\label{secwithoutstationary}
When (C2) is violated, the relation $\psi_1=\psi_2=\cdots=\psi_{T-1}$ might no longer hold. However, under (C1), (C3) and $H_0$, the marginal distribution function of $X_{0,t}$ can be well-approximated by some $\mathbb{F}$ on average. As a result, $\psi_t$'s can be well-approximated by some $\psi^*$ on average. Let $\mathbb{F}_t$ denote the distribution function of $X_{0,t}$. As long as the prediction error satisfies  
\begin{eqnarray*}
	\max_{1\le b\le B}  \frac{1}{T}\sum_{t=1}^{T}\int_{x}|\widehat{\psi}^{(-\ell)}(\nu_b|x)-\psi_t(\nu_b|x)|^2\mathbb{F}_t(dx)
	=O_p((nT)^{-c_0}),
\end{eqnarray*}
for some $c_0>1/2$, our test remains valid. 

\section{More on the OhioT1DM dataset}
\subsection{Detailed definitions of actions and rewards}\label{secAtRt}
We define $A_{0,t}$ as follows:

\begin{equation*}\label{discretization}
A_{0,t} =  \begin{cases}
0, & \hbox{In}_{0,t} = 0; \\
m, & 4m - 4 < \hbox{In}_{0,t} \le 4m \quad (m = 1, 2, 3);\\
4, & \hbox{In}_{0,t} > 12.
\end{cases}
\end{equation*}
The Index of Glycemic Control is chosen as the immediate reward $R_{0,t}$, defined by

\begin{equation*}\label{reward: IGC}
R_{0,t} =  \begin{cases}
- \frac{1}{30}(80 - \hbox{G}_{0,t + 1})^2, & G_{0,t + 1} < 80; \\
0, & 80 \le \hbox{G}_{0,t + 1} \le 140; \\
-\frac{1}{30}(\hbox{G}_{0,t + 1} - 140)^{1.35}, & 140 \le G_{0,t + 1}.
\end{cases}
\end{equation*}

\subsection{Detailed procedure for value evaluation in simulations}\label{supp:num_simu}

In Section \ref{sec:ohio_simu}, we compare the policies learned with the selected order $\widehat{\kappa}_0$ and fixed orders $k\in \{1,\cdots,10\}$. 
Below, we provide more details on computing the value $V^{(l)}(k)$. 

\begin{enumerate}
	\item In the $l$-th simulation, generate $N$ trajectories $\{(S_{j,t},A_{j,t})\}_{1 \le j \le N, 0\le t \le 1344}$, and apply Algorithm 2 with $\alpha=0.01$ and $K=10$ to estimate an order $\widehat{\kappa}_0^{(l)}$. Also generate 100 trajectories of length 10 with the model described in Section \ref{sec:ohio_simu}, denoted by 
	$\{(S^e_{j,t},A^e_{j,t})\}_{1 \le j \le 100, 0\le t < 10}$.
	\item For $k=1, \ldots, 10$,  apply FQI (see below) to the concatenated  data $\{(S_{j,t}(k),A_{j,t}(k), R_{j,t}(k))\}_{1\le j\le N,0\le t\le 1344-k}$ to learn an optimal policy $\widehat{\pi}^{(l)}(k)$. 
	
	\item For each initial trajectory $\{(S^e_{j,t},A^e_{j,t})\}_{0\le t < 10}$, generate the data $\{(S^e_{j,t},A^e_{j,t},R^e_{j,t})\}_{10\le t < 60}$ following $\widehat{\pi}^{(l)}(k)$. Compute the value $V^{(l)}(k)$ by
	\begin{eqnarray*}
		V^{(l)}(k)=\frac{1}{100}\sum_{j=1}^{100} \sum_{t=10}^{50} \gamma^{t-10} R_{j,t}^e,
	\end{eqnarray*}
	with $\gamma=0.9$. 
\end{enumerate}


\begin{algorithm}
	\caption{Fitted-Q iteration}\label{FQI}
	\begin{algorithmic}
		\STATE \textbf{Input:} Data $\{S_{j,t},A_{j,t},R_{j,t},S_{j,t+1}\}_{j,t}$,  function class $\mathcal{F}$, decay rate $\gamma$, action space $\mathcal{A}$
		\STATE Randomly pick $Q_0 \in \mathcal{F}$
		\FOR{$k = 1, \dots, K$} 
		
		\STATE Update target values $Z_{j,t} = R_{j,t} + \gamma \max_{a \in \mathcal{A}} Q_{k-1}(S_{j,t+1}, a)$ for all $(j,t)$;
		
		\STATE Solve a regression problem to update the $Q$-function:\\
		\quad \quad    $Q_k = \argmin_{Q \in \mathcal{F}} \frac{1}{n} \sum_{i=1}^n \{Q(S_{j,t}, A_{j,t}) - Z_{j,t}\}^2$
		\ENDFOR
		\STATE \textbf{Output:} The estimated optimal policy $\widehat{\pi}(\cdot) = \argmax_{a \in \mathcal{A}} Q_{K}(\cdot,a)$
	\end{algorithmic}	
\end{algorithm}	
In our experiment,we use random forests to estimate the Q function during each iteration. The number of trees are set as 100 and the other hyperparameters are selected by 5-fold cross-validation. The decay rate $\gamma$ is set to $0.9$.

\subsection{Detailed procedure for value evaluation in real data analysis}\label{supp:num_real}

In Section \ref{sec:real_ohio}, we compare policies learned by assuming the data follows a $k$-th order MDP for $k\in \{1,\cdots,10\}$. The policies are estimated by FQI. 
To evaluate the values of these policies based on the real dataset, we apply the Fitted-Q evaluation (FQE) algorithm. Similar to FQI, it is an iterative algorithm based on the Bellman equation. 
We recap the steps below.

\begin{algorithm}
	\caption{Fitted-Q evaluation}\label{FQE}
	\begin{algorithmic}
		\STATE \textbf{Input:} Data $\{S_{j,t},A_{j,t},R_{j,t},S_{j,t+1}\}_{j,t}$, policy $\pi$, function class $\mathcal{F}$, decay rate $\gamma$
		\STATE  Randomly pick $Q_0 \in \mathcal{F}$\;
		\FOR{$k = 1, \dots, K$} 
		
		\STATE Update target values $Z_{j,t} = R_{j,t} + \gamma Q_{k-1}(S_{j,t+1}, \pi(S_{j,t+1}))$ for all $(j,t)$;
		
		\STATE Solve a regression problem to update the $Q$-function:\\
		\quad \quad    $Q_k = \argmin_{Q \in \mathcal{F}} \frac{1}{n} \sum_{i=1}^n \{Q(S_{j,t}, A_{j,t}) - Z_{j,t}\}^2$
		\ENDFOR
		\STATE \textbf{Output:} The estimated value $\widehat{V}(\cdot)=Q_K(\cdot,\pi(\cdot))$
	\end{algorithmic}	
\end{algorithm}	

Denote the trajectories for the six patients in the OhioT1DM dataset by
$\{(S_{i,t},A_{i,t})\}_{1 \le i \le 6,1\le t \le 1100}$, and let the index set $\mathcal{I} = \{1,2,3,4,5,6\}$. We now describe the evaluation procedure in more details: 

\begin{enumerate}
	\item In $l=1, \ldots, 20$, divide $\mathcal{I}$ into a training set $\mathcal{D}^{(l)}_1$ and an validation set $\mathcal{D}^{(l)}_2 = (\mathcal{D}_1^{(l)})^c$ with $|\mathcal{D}^{(l)}_1|=|\mathcal{D}^{(l)}_2|=3$. 
	\item For each $l\in \{1, \ldots, 20\}$, $k\in \{1, \dots, 10\}$, apply FQI to the data $\{(S_{j, t}(k), A_{j, t}(k), R_{j, t}(k))\}_{j \in \mathcal{D}_1^{(l)}, 0\le t\le 1100-k+1}$ to learn an optimal policy $\widehat{\pi}^{(l)}(k)$. 
	\item For each $l\in \{1, \ldots, 20\}$, $k\in \{1, \dots, 10\}$, apply FQE to the data $\{(S_{j, t}(k), A_{j, t}(k), R_{j, t}(k))\}_{j \in \mathcal{D}_2^{(l)}, 0\le t\le 1100-k+1}$ to estimate the state-value function of  $\widehat{\pi}^{(l)}(k)$, denoted by $\widehat{V}^{(l)}_k(\cdot)$. Generate 100 trajectories of length 10 according to the simulation model in Section \ref{sec:ohio_simu}. Denote them by 
	$\{(S^e_{j,t},A^e_{j,t})\}_{1 \le j \le 100, 0\le t < 10}$. Calculate the value under $\widehat{\pi}^{(l)}(k)$ by
	\begin{eqnarray*}
		V^{(l)}(k) = \frac{1}{100} \sum_{j = 1}^{100} \widehat{V}_k^{(l)}(S^e_{j,(10-k)}(k)).
	\end{eqnarray*} 
	\item Average over the $20$ splits to compute the average value for each $k$ by $V(k) = \sum_{l = 1}^{20} V^{(l)}(k)/20$.
\end{enumerate}

For both FQI and FQE, we use random forests to estimate the regression function. The number of trees are set to 75 and the other hyperparameters are selected by 5-fold cross-validation. 
We set $\gamma = 0.9$ in our experiments. 

%
%
%
%
%
%
%


\section{Technical proofs}
\subsection{Proof of Lemma \ref{keylemma}}\label{secprooflemma1}
Consider a policy $\pi=\{\pi_t\}_{t\ge 0} \in \hbox{HR}$. Suppose there exists some $\{\pi_t^*\}_{t\ge 0}$ such that $\pi_t(\cdot|\bar{\bm{S}}_{0,t})=\pi_t^*(\cdot|S_{0,t})$ almost surely for any $t\ge 0$. We refer to such a policy $\pi$ as a \textit{Markov} policy. In addition, $\pi$ is a \textit{deterministic} policy if and only if $\pi_t(a|\bar{\bm{S}}_{0,t})\in \{0,1\}$ almost surely for any $t\ge 0$ and $a\in \mathcal{A}$. 
Let MR denotes the set of Markov policies and SD denote the set of deterministic stationary policies, we have $\hbox{SD}\subseteq \hbox{SR}\subseteq \hbox{MR}\subseteq \hbox{HR}$. In the following, we focus on proving
\begin{eqnarray*}
	\sup_{\pi \in \scriptsize{\hbox{HR}}}  V(\pi;s)=\sup_{\pi \in \scriptsize{\hbox{SD}}} V(\pi;s),\,\,\,\,\,\,\,\,\forall s\in \mathbb{S}.
\end{eqnarray*}
Since $\hbox{SD}\subseteq \hbox{SR}$, the assertion in Lemma \ref{keylemma} is thus satisfied. 

We begin by providing a sketch of the proof. Our proof is divided into three steps. In the first step, we show 
\begin{eqnarray*}
	\sup_{\pi \in \scriptsize{\hbox{HR}}}  V(\pi;s)=\sup_{\pi \in \scriptsize{\hbox{MR}}} V(\pi;s),\,\,\,\,\,\,\,\,\forall s\in \mathbb{S}.
\end{eqnarray*}
To prove this,
we show in Section \ref{secsometech1} that for any such $\pi\in \hbox{HR}$ and any $s$, there exists a Markov policy $\pi^*=\{\pi_t^*\}_{t\ge 0}$ where each $\pi_t^*$ depends on $S_{0,t}$ only such that 
\begin{eqnarray}\label{importanteq1}
\prob^{\pi}(A_{0,t}=a,S_{0,t}\in \mathcal{S}| S_{0,0}=s)=\prob^{\pi^*}(A_{0,t}=a,S_{0,t}\in \mathcal{S}|S_{0,0}=s),
\end{eqnarray}
for any $t\ge 0$, $a\in \mathcal{A}$, $\mathcal{S}\subseteq \mathbb{S}$ and $s\in \mathbb{S}$ where the probabilities $\prob^\pi$ and $\prob^{\pi^*}$ are taken by assuming the system dynamics follow $\pi$ and $\pi^*$, respectively. Under MA, we have
\begin{eqnarray*}
	\Mean^{\pi}(Y_{0,t}|S_{0,0}=s)=\Mean^{\pi} \{\Mean^{\pi}(Y_{0,t}|A_{0,t},S_{0,t},S_{0,0}=s)|S_{0,0}=x\}=\Mean^{\pi} \{r(A_{0,t},S_{0,t})|S_{0,0}=x\},
\end{eqnarray*}
for some function $r$. This together with \eqref{importanteq1} yields that
\begin{eqnarray*}
	\Mean^{\pi}(Y_{0,t}|S_{0,0}=s)=\Mean^{\pi^*}(Y_{0,t}|S_{0,0}=s),\,\,\,\,\,\,\,\,\forall t\ge 0,
\end{eqnarray*}
and hence $V(\pi;s)=V(\pi^*;s)$. This completes the proof for the first step. 

With a slight abuse of notation, for any $\pi \in \hbox{SD}$, we denote by $\pi(s)$ the action that the agent chooses according to $\pi$, given that the current state equals $s$. In the second step, we show for any bounded function $\nu(\cdot)$ on $\mathbb{S}$ that satisfies the optimal Bellman equation
\begin{eqnarray*}
	\nu(s)=\sup_{\pi \in \scriptsize{\hbox{SD}} } \left\{ r(\pi(s),s)+\gamma \int_{s'} \nu(s') \mathcal{P}(ds';\pi(s),s) \right\},\,\,\,\,\,\,\,\,\forall s\in \mathbb{S},
\end{eqnarray*} 
it satisfies
\begin{eqnarray}\label{importanteq2}
\nu(s)=\sup_{\pi^*\in \scriptsize{\hbox{MR}}} V(\pi^*;s),\,\,\,\,\,\,\,\,\forall s\in \mathbb{S}.
\end{eqnarray}
The proof of \eqref{importanteq2} is given in Section \ref{secsometech2}. 

For any function $\nu$, define the norm $\| \nu \|_{\infty}=\sup_{s\in \mathbb{S}} |\nu(s)|$. We have for any $\nu_1$ and $\nu_2$ that
\begin{eqnarray*}
	\sup_{x} \left|\sup_{\pi \in \scriptsize{\hbox{SD}} } \left\{ r(\pi(s),s)+\gamma \int_{s'} \nu_1(s') \mathcal{P}(ds';\pi(s),s)\right\} -\sup_{\pi \in \scriptsize{\hbox{SD}} } \left\{ r(\pi(s),s)+\gamma \int_{s'} \nu_2(s') \mathcal{P}(ds';\pi(s),s) \right\}\right|\\ \le \gamma\sup_{\pi \in \scriptsize{\hbox{SD}}} \sup_{s\in \mathbb{S}}  \left|\int_{s'} \nu_1(s') \mathcal{P}(ds';\pi(s),s) -\int_{s'} \nu_2(s') \mathcal{P}(ds';\pi(s),s) \right|\\
	\le \gamma\sup_{\pi \in \scriptsize{\hbox{SD}}} \sup_{s\in \mathbb{S}}  \left|\int_{s'} \|\nu_1-\nu_2\|_{\infty} \mathcal{P}(ds';\pi(s),s) \right|\le \gamma \|\nu_1-\nu_2\|_{\infty}.
\end{eqnarray*}
By Banach's fix point theorem, there exists a unique value function $\nu_0$ that satisfies the optimal Bellman equation. Combining this together with the results obtained in the first two steps, we obtain that $\nu_0$ satisfies $\nu_0(s)=\sup_{\pi \in \scriptsize{\hbox{HR}}}  V(\pi;s)$ for any $s\in \mathbb{S}$. The proof is thus completed if we can show there exists a deterministic stationary policy $\pi^{**}$ that satisfies
\begin{eqnarray}\label{importanteq3}
\nu_0(s)=V(\pi^{**};s),\,\,\,\,\,\,\,\,\forall s\in \mathbb{S}.
\end{eqnarray}
We put the proof of \eqref{importanteq3} in Section \ref{secproofimportanteq3}. 

\subsubsection{Proof of \eqref{importanteq1}}\label{secsometech1}
Apparently, \eqref{importanteq1} holds with $t=0$. Suppose \eqref{importanteq1} holds for $t=k$. We now show \eqref{importanteq1} holds for $t=k+1$. Under MA, we have
\begin{eqnarray*}
	&&\prob^{\pi}(S_{0,k+1}\in \mathcal{S}|S_{0,0}=s)=\Mean^{\pi} \{\prob^{\pi}(S_{0,t+1}\in \mathcal{S}|A_{0,t},S_{0,t},S_{0,0}=s)|S_{0,0}=x\}\\
	&=&\Mean^{\pi} \{\mathcal{P}(\mathcal{S};A_{0,t},S_{0,t})|S_{0,0}=x\}
	=\Mean^{\pi^*} \{\mathcal{P}(\mathcal{S};A_{0,t},S_{0,t})|S_{0,0}=x\}=\prob^{\pi^*}(S_{0,k+1}\in \mathcal{S}|S_{0,0}=s)\stackrel{\Delta}{=}\mathbb{G}_{k+1}(\mathcal{S};s).
\end{eqnarray*}
Set $\pi^*_{k+1}$ to be the decision rule that satisfies
\begin{eqnarray*}
	\prob^{\pi}(A_{0,k+1}=a|S_{0,k+1},S_{0,0}=s)=\prob^{\pi_{k+1}^*}(A_{0,k+1}=a|S_{0,k+1}),\,\,\,\,\,\,\,\,\forall a\in \mathcal{A},
\end{eqnarray*}
it follows that
\begin{eqnarray*}
	\prob^{\pi}(A_{0,k+1}=a,S_{0,k+1}\in \mathcal{S}|S_{0,0}=s)=\int_{s'} \prob^{\pi}(A_{0,k+1}=a|S_{0,k+1}=s',S_{0,0}=s)\mathbb{G}_{k+1}(ds';s)\\
	=\int_{s'} \prob^{\pi^*}(A_{0,k+1}=a|S_{0,k+1}=s',S_{0,0}=s)\mathbb{G}_{k+1}(ds';s)=\prob^{\pi^*}(A_{0,k+1}=a,S_{0,k+1}\in \mathcal{S}|S_{0,0}=s).
\end{eqnarray*}
Thus, \eqref{importanteq1} holds for $t=k+1$ as well. The proof is hence completed. 

\subsubsection{Proof of \eqref{importanteq2}}\label{secsometech2}
We first show for any bounded function $\nu$ that satisfies
\begin{eqnarray}\label{cond1}
\nu(s)\ge \sup_{\pi \in \scriptsize{\hbox{SD}} } \left\{ r(\pi(s),s)+\gamma \int_{s'} \nu(s') \mathcal{P}(ds';\pi(s),s) \right\},\,\,\,\,\,\,\,\,\forall s\in \mathbb{S},
\end{eqnarray} 
we have 
\begin{eqnarray}\label{importanteq2.3}
\nu(s)\ge \sup_{\pi^*\in \scriptsize{\hbox{MR}}} V(\pi^*;s),\,\,\,\,\,\,\,\,\forall s\in \mathbb{S}.
\end{eqnarray}
Then, we show for any bounded function $\nu$ that satisfies
\begin{eqnarray*}
	\sup_{s\in \mathbb{S}}\left[\nu(s)-\sup_{\pi \in \scriptsize{\hbox{SD}} } \left\{ r(\pi(s),s)+\gamma \int_{s'} \nu(s') \mathcal{P}(ds';\pi(s),s) \right\}\right]\le 0,
\end{eqnarray*} 
we have 
\begin{eqnarray}\label{importanteq2.6}
\nu(s)\le \sup_{\pi^*\in \scriptsize{\hbox{MR}}} V(\pi^*;s),\,\,\,\,\,\,\,\,\forall s\in \mathbb{S}.
\end{eqnarray}
The proof is hence completed. 

\textit{Proof of \eqref{importanteq2.3}: }
Consider an arbitrary deterministic Markov policy $\pi^*=\{\pi^*_t\}_{t\ge 0}$. With a slight abuse of notation, we denote by $\pi^*_t(s)$ the action that the agent chooses following $\pi^*_t$, given that the current state equals $s$. 
It follows from \eqref{cond1} that
\begin{eqnarray*}
	\nu(s)\ge r(\pi_0^*(s),s)+\gamma \int_{s'} \nu(s') \mathcal{P}(ds';\pi_0^*(s),s),\,\,\,\,\,\,\,\,\forall s\in \mathbb{S}.
\end{eqnarray*}
By iteratively applying \eqref{cond1}, we have
\begin{eqnarray*}
	\nu(s)\ge r(\pi_0^*(s),s)+\sum_{k=1}^K \gamma^k \Mean^{\pi^*} \{r(A_{0,k},X_{0,k})|S_{0,0}=x\}
	+\gamma^{K+1} \Mean^{\pi^*} \{\nu(X_{0,K+1})|S_{0,0}=x\},\,\,\,\,\,\,\,\,\forall s\in \mathbb{S}. 
\end{eqnarray*}
Since $\nu$ is bounded, the last term on the right-hand-side (RHS) converges to zero uniformly in $x$, as $K\to \infty$. Let $K\to \infty$, we obtain $\nu(s)\ge V(\pi^*;s)$, for any $s\in \mathbb{S}$ and any deterministic Markov policy $\pi^*$. Using Lemma 4.3.1 of \citet{Puterman1994}, we can similarly show $\nu(s)\ge V(\pi^*;s)$ for any $s\in \mathbb{S}$ and $\pi^*\in \hbox{MR}$. This completes the proof of \eqref{importanteq2.3}.

\textit{Proof of \eqref{importanteq2.6}: }By definition, we have
\begin{eqnarray*}
	\inf_{\pi \in \scriptsize{\hbox{SD}} }\sup_{s\in \mathbb{S}}\left[\nu(s)- \left\{ r(\pi(s),s)+\gamma \int_{s'} \nu(s') \mathcal{P}(ds';\pi(s),s) \right\}\right]\le 0.
\end{eqnarray*}
Thus, for any $\epsilon>0$, there exists some $\pi_0\in$ SD that satisfies
\begin{eqnarray}\label{importanteq2.7}
\sup_{s\in \mathbb{S}}\left[\nu(s)- \left\{ r(\pi_0(s),s)+\gamma \int_{s'} \nu(s') \mathcal{P}(ds';\pi_0(s),s) \right\}\right]\le \epsilon.
\end{eqnarray}
Consider the following bounded linear operator $\mathcal{L}_0$,
\begin{eqnarray*}
	\mathcal{L}_0 \nu(s)= \int_{s'} \nu(s') \mathcal{P}(ds';\pi_0(s),s),
\end{eqnarray*}
defined on the space of bounded functions. Let $\mathcal{I}_0$ denote the identity operator. Since $\gamma<1$, the operator $\mathcal{I}_0-\gamma \mathcal{L}_0$ is invertible and its inverse equals $\sum_{k=0}^{+\infty} \gamma^k \mathcal{L}_0^k$. It follows from \eqref{importanteq2.7} that
\begin{eqnarray*}
	\nu(s)\le \sum_{k=0}^{+\infty} \gamma^k \mathcal{L}_0^k \{r(\pi_0(s),s)+\epsilon\},\,\,\,\,\,\,\,\,\forall s\in \mathbb{S}.
\end{eqnarray*}
Since $V(\pi_0;s)=\sum_{k=0}^{+\infty} \gamma^k \mathcal{L}_0^k r(\pi_0(s),s)$ and $\sum_{k=0}^{+\infty} \gamma^k \mathcal{L}_0^k \epsilon\le \epsilon/(1-\gamma)$, we obtain
\begin{eqnarray*}
	\nu(s)\le V(\pi_0;s)+\frac{\epsilon}{1-\gamma}.
\end{eqnarray*}
Let $\epsilon\to 0$, we obtain $\nu(s)\le \sup_{\pi^*\in \scriptsize{\hbox{MR}}} V(\pi^*;s)$ for any $x$. The proof is hence completed. 

\subsubsection{Proof of \eqref{importanteq3}}\label{secproofimportanteq3}
Since $\nu_0(\cdot)$ satisfies the optimal Bellman equation, we have
\begin{eqnarray*}
	\nu_0(s)=\argmax_{\pi\in \scriptsize{\hbox{SD}}}\left\{ r(\pi(s),s)+\gamma \int_{s'} \nu_0(s') \mathcal{P}(ds';\pi(s),s) \right\}.
\end{eqnarray*}
Let $\mathcal{A}_s$ be the available set of actions at a given state $s$. 
As a result, we have 
\begin{eqnarray*}
	\nu_0(s)=\argmax_{a \in \mathcal{A}_s}\left\{ r(a,s)+\gamma \int_{s'} \nu_0(s') \mathcal{P}(ds';a,s) \right\}.
\end{eqnarray*}
Since $\mathcal{A}$ is finite, so is $\mathcal{A}_s$. As a result, the above argmax is achievable. Let $\pi^{**}(s)$ be the action such that the above argmax is achieved, we have 
\begin{eqnarray*}
	\nu_0(s)= r(\pi^{**}(s),s)+\gamma \int_{s'} \nu_0(s') \mathcal{P}(ds';\pi^{**}(s),s).
\end{eqnarray*}
Similar to the proof of \eqref{importanteq2}, we can show $\nu_0(s)=V(\pi^{**};s)$, for all $s\in \mathbb{S}$. The proof is hence completed. 

\subsection{Proof of Theorem \ref{lemma3}} 
The proof is divided into two parts. In the first part, we show \eqref{CIA} $\Rightarrow$ \eqref{CIA2}. In the second part, we show \eqref{CIA2} $\Rightarrow$ \eqref{CIA}.

\subsubsection{Part 1}
Under \eqref{CIA}, $S_{0,t+q+1}\independent \{X_{0,j}\}_{j< t+q} | X_{0,t+q}$. It follows that
\begin{eqnarray*}
	\Mean [\exp(i\mu^\top S_{0,t+q+1}+i\nu^\top X_{0,t-1})|\{X_{0,j}\}_{t\le j\le t+q}]
	=\varphi_{t+q}(\mu|X_{0,t+q}) \Mean [(i\nu^\top X_{0,t-1})|\{X_{0,j}\}_{t\le j\le t+q}].
\end{eqnarray*}
The proof is hence completed. 

\subsubsection{Part 2}
We introduce the following lemma before presenting the proof. 
\begin{lemma}\label{lemmaS1}
	For any random vectors $Z_1\in\mathbb{R}^{q_1},Z_2\in\mathbb{R}^{q_2},Z_3\in\mathbb{R}^{q_3}$, suppose $\Mean \{\exp(i\mu_1^\top Z_1)|Z_3\} \Mean \{\exp(i\mu_2^\top Z_2)|Z_3\}=\Mean \{\exp(i\mu_1^\top Z_1+i\mu_2^\top Z_2)|Z_3\}$ for any $\mu_1\in \mathbb{R}^{q_1}$, $\mu_2\in \mathbb{R}^{q_2}$ almost surely. Then we have $Z_1\independent Z_2|Z_3$.   
\end{lemma}
Let $q=0$. By \eqref{CIA2}, we obtain
\begin{eqnarray*}
	\varphi_{t}(\mu|X_{0,t})\Mean\{\exp(i\nu^\top X_{0,t-1})|X_{0,t}\}=\Mean [\exp(i\mu^\top S_{0,t+1}+i\nu^\top X_{0,t-1})|X_{0,t}],
\end{eqnarray*}
for any $t>0$, $\mu\in \mathbb{R}^p$, $\nu\in \mathbb{R}^{p+1}$. By Lemma \ref{lemmaS1}, we obtain
\begin{eqnarray}\label{independence0}
S_{0,t+1}\independent X_{0,t-1}|X_{0,t},\,\,\,\,\,\,\,\,\forall t>0.
\end{eqnarray}
Set $q=1$, we have by \eqref{CIA2} that
\begin{eqnarray}\label{independence05}
\varphi_{t+1}(\mu|X_{0,t+1})\Mean\{\exp(i\nu^\top X_{0,t-1})|X_{0,t},X_{0,t+1}\}=\Mean [\exp(i\mu^\top S_{0,t+2}+i\nu^\top X_{0,t-1})|X_{0,t}, X_{0,t+1}],
\end{eqnarray}
for any $t>0$, $\mu\in \mathbb{R}^p$, $\nu\in \mathbb{R}^{p+1}$. 
For any $v\in\mathbb{R}^{p+1}$, multiply both sides of \eqref{independence05} by $\exp(iv^\top X_{0,t})$ and take expectation with respect to $X_{0,t}$ conditional on $X_{0,t+1}$, we obtain
\begin{eqnarray*}
	\Mean \{\exp(i\mu^\top S_{0,t+2})|X_{0,t+1}\} \Mean \{\exp(iv^\top X_{0,t-1}+ i\nu^\top X_{0,t})|X_{0,t+1}\}
	=\Mean [\exp(i\mu^\top S_{0,t+2}+iv^\top X_{0,t-1}+i\nu^\top X_{0,t})|X_{0,t+1}].
\end{eqnarray*}
By Lemma \ref{lemmaS1}, we obtain
\begin{eqnarray}\label{independence1}
S_{0,t+2}\independent X_{0,t-1},X_{0,t}|X_{0,t+1},\,\,\,\,\,\,\,\,\forall t>0.
\end{eqnarray}
Similarly, we can show
\begin{eqnarray}\label{independence2}
S_{0,t+q+1}\independent \{S_{0,j}\}_{t-1 \le j<t+q}|X_{0,t+q},\,\,\,\,\,\,\,\,\forall t.
\end{eqnarray}
Combining \eqref{independence0} with \eqref{independence1} and \eqref{independence2} yields \eqref{CIA}. The proof is hence completed. 

\subsubsection{Proof of Lemma \ref{lemmaS1}}
Let $\widetilde{Z}_1,\widetilde{Z}_2$ be independent copies of $Z_1,Z_2$ such that $\widetilde{Z}_1|Z_3\stackrel{d}{=}Z_1|Z_3$, $\widetilde{Z}_2|Z_3\stackrel{d}{=}Z_2|Z_3$ and that $\widetilde{Z}_1\independent \widetilde{Z}_2 |\widetilde{Z}_3$. Consider any $\mu_1\in \mathbb{R}^{q_1},\mu_2\in \mathbb{R}^{q_2},\mu_3\in \mathbb{R}^{q_3}$, we have
\begin{eqnarray}\label{part2eq1}
&&\Mean \exp(i\mu_1^\top \widetilde{Z}_1+i\mu_2^\top \widetilde{Z}_2+i\mu_3^\top Z_3)=\Mean[\exp(i\mu_3^\top Z_3) \Mean \{\exp(i\mu_1^\top \widetilde{Z}_1+i\mu_2^\top \widetilde{Z}_2)|Z_3\}]\\\nonumber
&=&\Mean[\exp(i\mu_3^\top Z_3)\Mean \{\exp(i\mu_1^\top \widetilde{Z}_1)|Z_3\} \Mean \{\exp(i\mu_2^\top \widetilde{Z}_2)|Z_3 \}]=\Mean[\exp(i\mu_3^\top Z_3)\Mean \{\exp(i\mu_1^\top Z_1)|Z_3\} \Mean \{\exp(i\mu_2^\top Z_2)|Z_3 \}].
\end{eqnarray} 
Under the condition in Lemma \ref{lemmaS1}, we have
\begin{eqnarray*}
	\Mean[\exp(i\mu_3^\top Z_3)\Mean \{\exp(i\mu_1^\top Z_1)|Z_3\} \Mean \{\exp(i\mu_2^\top Z_2)|Z_3 \}]=\Mean[\exp(i\mu_3^\top Z_3)\Mean \{\exp(i\mu_1^\top Z_1+i\mu_2^\top Z_2)|Z_3 \}]\\
	=\Mean \exp(i\mu_1^\top Z_1+i\mu_2^\top Z_2+i\mu_3^\top Z_3).
\end{eqnarray*}
This together with \eqref{part2eq1} yields 
\begin{eqnarray*}
	\Mean \exp(i\mu_1^\top \widetilde{Z}_1+i\mu_2^\top \widetilde{Z}_2+i\mu_3^\top Z_3)=\Mean \exp(i\mu_1^\top Z_1+i\mu_2^\top Z_2+i\mu_3^\top Z_3).
\end{eqnarray*}
As a result, $(\widetilde{Z}_1,\widetilde{Z}_2,Z_3)$ and $(Z_1,Z_2,Z_3)$ have same characteristic functions. Therefore, we have  $(\widetilde{Z}_1,\widetilde{Z}_2,Z_3)\stackrel{d}{=}(Z_1,Z_2,Z_3)$. By construction, we have $\widetilde{Z}_1 \independent \widetilde{Z}_2 |Z_3$. It follows that $Z_1\independent Z_2|Z_3$. 

\subsection{Proof of Theorem \ref{thm1}}
We focus on proving Theorem \ref{thm1} in the more challenging setting where $T\to \infty$. The number of trajectories $n$ can be either bounded or growing to $\infty$. The case where $T$ is bounded can be proven using similar arguments. We begin by providing an outline of the proof. 
For any $q,\mu,\nu$, define
\begin{eqnarray*}
	\Gamma^*(q,\mu,\nu)=\frac{1}{n(T-q-1)} \sum_{j=1}^{n} \sum_{t=1}^{T-q-1}\{\exp(i\mu^\top S_{j,t+q+1})
	-\varphi^{*}(\mu|X_{j,t+q}) \} \{\exp(i\nu^\top X_{j,t-1})-\psi^*(\nu|X_{j,t})\}.
\end{eqnarray*}
Denote by $\Gamma^*_R$ and $\Gamma^*_I$ the real and imaginary part of $\Gamma^*$, respectively. 

We break the proof into three steps. In the first step, we show
\begin{eqnarray}\label{eqstep1}
\max_{b\in \{1,\cdots,B\}}\max_{q\in \{0,\cdots,Q\}} \sqrt{n(T-q-1)} |\widehat{\Gamma}(q,\mu_b,\nu_b)-\Gamma^*(q,\mu_b,\nu_b)|=o_p(\log^{-1/2} (nT)).
\end{eqnarray}
Proof of \eqref{eqstep1} relies largely on Condition (C4) which requires $\widehat{\varphi}$ and $\widehat{\psi}$ to satisfy certain uniform convergence rates. This further implies that 
\begin{eqnarray}\label{eqstep1star}
\widehat{S}=S^*+o_p(\log^{-1/2} (nT)),
\end{eqnarray}
where 
\begin{eqnarray*}
	S^*=\max_{b\in \{1,\cdots,B\}}\max_{q\in \{0,\cdots,Q\}} \sqrt{n(T-q-1)}\max(|\Gamma_R^*(q,\mu_b,\nu_b)|, |\Gamma_I^*(q,\mu_b,\nu_b)|).
\end{eqnarray*}
In the second step, we show for any $z\in \mathbb{R}$ and any sufficiently small $\varepsilon>0$,
\begin{eqnarray*}
	\prob(S^*\le z)\ge \prob(\|N(0,V_0)\|_{\infty}\le z-\varepsilon\log^{-1/2} (nT))-o(1),\\
	\prob(S^*\le z)\le \prob(\|N(0,V_0)\|_{\infty}\le z+\varepsilon\log^{-1/2} (nT))+o(1),
\end{eqnarray*}
where the matrix $V_0$ is defined in Step 2 of the proof. This together with \eqref{eqstep1star} yields that
\begin{eqnarray}\label{eqstep2}
\prob(\widehat{S}\le z)\ge \prob(\|N(0,V_0)\|_{\infty}\le z-2\varepsilon\log^{-1/2} (nT))-o(1),\\\label{eqstep3}
\prob(\widehat{S}\le z)\le \prob(\|N(0,V_0)\|_{\infty}\le z+2\varepsilon\log^{-1/2} (nT))+o(1).
\end{eqnarray}
The proposed Bootstrap algorithm repeatedly generate random variables from $\|N(0,\widehat{V})\|_{\infty}$ where the detailed form of $\widehat{V}$ is given in the third step of the proof. The critical values $\widehat{c}_{\alpha}$ is chosen to be the upper $\alpha$-th quantile of $\|N(0,\widehat{V})\|_{\infty}$. In the third step, we show $\|V_0-\widehat{V}\|_{\infty,\infty}=O((nT)^{-c^{**}})$ for some $c^{**}>0$ with probability tending to $1$, where $\|\cdot\|_{\infty,\infty}$ denotes the elementwise max-norm. Combining this upper bound with some arguments used in proving  \eqref{eqstep2} and \eqref{eqstep3}, we can show with probability tending to $1$ that
\begin{eqnarray*}
	\prob(\widehat{S}\le z)\ge \prob(\|N(0,\widehat{V})\|_{\infty}\le z-2\varepsilon\log^{-1/2} (nT)|\widehat{V})-o(1),\\
	\prob(\widehat{S}\le z)\le \prob(\|N(0,\widehat{V})\|_{\infty}\le z+2\varepsilon\log^{-1/2} (nT)|\widehat{V})+o(1), 
\end{eqnarray*}
for any sufficiently small $\varepsilon>0$ where $\prob(\cdot|\widehat{V})$ denotes the conditional probability given $\widehat{V}$. Set $z=\widehat{c}_{\alpha}$. It follows from that
\begin{eqnarray}\label{eqstep3.1}
\prob(\widehat{S}\le \widehat{c}_{\alpha})\ge \prob(\|N(0,\widehat{V})\|_{\infty}\le \widehat{c}_{\alpha}-2\varepsilon\log^{-1/2} (nT)|\widehat{V})-o(1),\\\label{eqstep3.2}
\prob(\widehat{S}\le \widehat{c}_{\alpha})\le \prob(\|N(0,\widehat{V})\|_{\infty}\le \widehat{c}_{\alpha}+2\varepsilon\log^{-1/2} (nT)|\widehat{V})+o(1),
\end{eqnarray}
with probability tending to $1$. 
Under the given conditions in Theorem \ref{thm1}, the diagonal elements in $V_0$ are bounded away from zero. With probability tending to $1$, the diagonal elements in $\widehat{V}$ is bounded away from zero as well. It follows from Theorem 1 of \citet{chernozhukov2017detailed} that conditional on $\widehat{V}$, 
\begin{eqnarray*}
	\prob(\|N(0,\widehat{V})\|_{\infty}\le \widehat{c}_{\alpha}+2\varepsilon\log^{-1/2} (nT)|\widehat{V})-\prob(\|N(0,\widehat{V})\|_{\infty}\le \widehat{c}_{\alpha}-2\varepsilon\log^{-1/2} (nT)|\widehat{V})\\
	\le O(1) \varepsilon \log^{1/2}(BQ) \log^{-1/2}(nT),
\end{eqnarray*}
with probability tending to $1$, where $O(1)$ denotes some positive constant that is independent of $\varepsilon$. Under the given conditions on $B$ and $Q$, we obtain with probability tending to $1$ that, 
\begin{eqnarray*}
	\prob(\|N(0,\widehat{V})\|_{\infty}\le \widehat{c}_{\alpha}+2\varepsilon\log^{-1/2} (nT)|\widehat{V})-\prob(\|N(0,\widehat{V})\|_{\infty}\le \widehat{c}_{\alpha}-2\varepsilon\log^{-1/2} (nT)|\widehat{V})\le C^*\varepsilon,
\end{eqnarray*}
for some constant $C^*>0$. This together with \eqref{eqstep3.1} and \eqref{eqstep3.2} yields 
\begin{eqnarray*}
	|\prob(\widehat{S}\le \widehat{c}_{\alpha})-\prob(\|N(0,\widehat{V})\|_{\infty}\le \widehat{c}_{\alpha}|\widehat{V})|\le C^*\varepsilon+o(1),
\end{eqnarray*}
with probability tending to $1$. Notice that $\varepsilon$ can be made arbitrarily small. The validity of our test thus follows. 

In the following, we present our proof for each of the step. Suppose $\{\mu_b,\nu_b\}_{1\le b\le B}$ are fixed throughout the proof. 
Denote by $\widehat{\varphi}_R^{(\ell)}$, $\widehat{\varphi}_I^{(\ell)}$ the real and imaginary part of $\widehat{\varphi}^{(\ell)}$ respectively. Without loss of generality, we assume the absolute values of $\widehat{\varphi}_R^{(\ell)}$, $\widehat{\varphi}_I^{(\ell)}$ are uniformly bounded by $1$. 
\subsubsection{Step 1}
With some calculations, we can show that for any $q,\mu,\nu$,
\begin{eqnarray*}
	\widehat{\Gamma}(q,\mu,\nu)=\Gamma^*(q,\mu,\nu)+R_1(q,\mu,\nu)+R_2(q,\mu,\nu)+R_3(q,\mu,\nu),
\end{eqnarray*}
where the remainder terms $R_1,R_2$ and $R_3$ are given by
\begin{eqnarray*}
	R_1(q,\mu,\nu)&=&\frac{1}{n(T-q-1)} \sum_{\ell=1}^{\mathbb{L}}\sum_{j\in \mathcal{I}^{(\ell)}} \sum_{t=1}^{T-q-1}\{
	\varphi^{*}(\mu|X_{j,t+q})-\widehat{\varphi}^{(-\ell)}(\mu|X_{j,t+q}) \}\{\psi^*(\nu|X_{j,t})-\widehat{\psi}^{(-\ell)}(\nu|X_{j,t})\},\\
	R_2(q,\mu,\nu)&=&\frac{1}{n(T-q-1)} \sum_{\ell=1}^{\mathbb{L}}\sum_{j\in \mathcal{I}^{(\ell)}} \sum_{t=1}^{T-q-1}\{
	\exp(i\mu^\top S_{j,t+q+1})
	-\varphi^{*}(\mu|X_{j,t+q}) \}\{\psi^*(\nu|X_{j,t})-\widehat{\psi}^{(-\ell)}(\nu|X_{j,t})\},\\
	R_3(q,\mu,\nu)&=&\frac{1}{n(T-q-1)} \sum_{\ell=1}^{\mathbb{L}}\sum_{j\in \mathcal{I}^{(\ell)}} \sum_{t=1}^{T-q-1}\{
	\varphi^{*}(\mu|X_{j,t+q})-\widehat{\varphi}^{(-\ell)}(\mu|X_{j,t+q}) \}\{\exp(i\nu^\top X_{j,t-1})-\psi^*(\nu|X_{j,t})\}.
\end{eqnarray*}
It suffices to show 
\begin{eqnarray}\label{eqstep1eq1}
\max_{b\in \{1,\cdots,B\}}\max_{q\in \{0,\cdots,Q\}} \sqrt{n(T-q-1)} |R_m(q,\mu_b,\nu_b)|=o_p(\log^{-1/2} (nT)),
\end{eqnarray}
for $m=1,2,3$. In the following, we show \eqref{eqstep1eq1} holds with $m=1,2$. Using similar arguments, one can show \eqref{eqstep1eq1} holds with $m=3$. 

\textbf{Proof of \eqref{eqstep1eq1} with $m=1$: }Since $\mathbb{L}$ is fixed, it suffices to show 
\begin{eqnarray}\label{eqstep1eq2}
\max_{b\in \{1,\cdots,B\}} \max_{q\in \{0,\cdots,Q\}} \sqrt{n(T-q-1)}|R_{1,\ell}(q,\mu_b,\nu_b)|=o_p(\log^{-1/2} (nT)),
\end{eqnarray}
where $R_{1,\ell}(q,\mu_b,\nu_b)$ is defined by 
\begin{eqnarray*}
	\frac{1}{n(T-q-1)} \sum_{j\in \mathcal{I}^{(\ell)}} \sum_{t=1}^{T-q-1}\{
	\varphi^{*}(\mu_b|X_{j,t+q})-\widehat{\varphi}^{(-\ell)}(\mu_b|X_{j,t+q}) \}\{\psi^*(\nu_b|X_{j,t})-\widehat{\psi}^{(-\ell)}(\nu_b|X_{j,t})\}.
\end{eqnarray*}
Similarly, let $\varphi^*_R$ and $\varphi^*_I$ denote the real and imaginary part of $\varphi^*$. We can rewrite $R_{1,\ell}(q,\mu_b,\nu_b)$ as $R_{1,\ell}^{(1)}(q,\mu_b,\nu_b)-R_{1,\ell}^{(2)}(q,\mu_b,\nu_b)+iR_{1,\ell}^{(3)}(q,\mu_b,\nu_b)+iR_{1,\ell}^{(4)}(q,\mu_b,\nu_b)$ where
\begin{eqnarray*}
	R_{1,\ell}^{(1)}(q,\mu_b,\nu_b)=\frac{1}{n(T-q-1)} \sum_{j\in \mathcal{I}^{(\ell)}} \sum_{t=1}^{T-q-1}\{
	\varphi^{*}_R(\mu_b|X_{j,t+q})-\widehat{\varphi}^{(-\ell)}_R(\mu_b|X_{j,t+q}) \}\{\psi^*_R(\nu_b|X_{j,t})-\widehat{\psi}^{(-\ell)}_R(\nu_b|X_{j,t})\},\\
	R_{1,\ell}^{(2)}(q,\mu_b,\nu_b)=\frac{1}{n(T-q-1)} \sum_{j\in \mathcal{I}^{(\ell)}} \sum_{t=1}^{T-q-1}\{
	\varphi^{*}_I(\mu_b|X_{j,t+q})-\widehat{\varphi}^{(-\ell)}_I(\mu_b|X_{j,t+q}) \}\{\psi^*_I(\nu_b|X_{j,t})-\widehat{\psi}^{(-\ell)}_I(\nu_b|X_{j,t})\},\\
	R_{1,\ell}^{(3)}(q,\mu_b,\nu_b)=\frac{1}{n(T-q-1)} \sum_{j\in \mathcal{I}^{(\ell)}} \sum_{t=1}^{T-q-1}\{
	\varphi^{*}_R(\mu_b|X_{j,t+q})-\widehat{\varphi}^{(-\ell)}_R(\mu_b|X_{j,t+q}) \}\{\psi^*_I(\nu_b|X_{j,t})-\widehat{\psi}^{(-\ell)}_I(\nu_b|X_{j,t})\},\\
	R_{1,\ell}^{(4)}(q,\mu_b,\nu_b)=\frac{1}{n(T-q-1)} \sum_{j\in \mathcal{I}^{(\ell)}} \sum_{t=1}^{T-q-1}\{
	\varphi^{*}_R(\mu_b|X_{j,t+q})-\widehat{\varphi}^{(-\ell)}_R(\mu_b|X_{j,t+q}) \}\{\psi^*_I(\nu_b|X_{j,t})-\widehat{\psi}^{(-\ell)}_I(\nu_b|X_{j,t})\}.
\end{eqnarray*}
To prove \eqref{eqstep1eq2}, it suffices to show 
\begin{eqnarray}\label{eqstep1eq3}
\max_{b\in \{1,\cdots,B\}} \max_{q\in \{0,\cdots,Q\}} \sqrt{n(T-q-1)}|R_{1,\ell}^{(s)}(q,\mu_b,\nu_b)|=o_p(\log^{-1/2} (nT)),
\end{eqnarray}
for $s=1,2,3,4$. For brevity, we only show \eqref{eqstep1eq3} holds with $s=1$. 

By the Cauchy-Schwarz inequality, it suffices to show
\begin{eqnarray}\label{eqstep1eq4}
\max_{b\in \{1,\cdots,B\}} \max_{q\in \{0,\cdots,Q\}}\frac{1}{\sqrt{n(T-q-1)}} \sum_{j\in \mathcal{I}^{(\ell)}} \sum_{t=1}^{T}\{
\varphi^{*}_R(\mu_b|X_{j,t})-\widehat{\varphi}^{(-\ell)}_R(\mu_b|X_{j,t}) \}^2=o_p(\log^{-1/2} (nT)),\\ \label{eqstep1eq5}
\max_{b\in \{1,\cdots,B\}} \max_{q\in \{0,\cdots,Q\}}\frac{1}{\sqrt{n(T-q-1)}} \sum_{j\in \mathcal{I}^{(\ell)}} \sum_{t=1}^{T}\{\psi^*_R(\nu_b|X_{j,t})-\widehat{\psi}^{(-\ell)}_R(\nu_b|X_{j,t})\}^2=o_p(\log^{-1/2} (nT)).
\end{eqnarray}
In the following, we focus on proving \eqref{eqstep1eq4}. Proof of \eqref{eqstep1eq5} is similar and is thus omitted.

%
%
%
Under (C2) and (C3), it follows from Theorem 3.7 of \citet{Brad2005} that $\{X_{0,t}\}_{t\ge 0}$ is exponentially $\beta$-mixing, that is, the $\beta$-mixing coefficient of $\{X_{0,t}\}_{t\ge 0}$ $\beta_0(\cdot)$ satisfies $\beta_0(t)=O(\rho^t)$ for some $\rho<1$ and any $t\ge 0$. Let $n_0=|\mathcal{I}^{(\ell)}|=n/\mathbb{L}$ and suppose $\mathcal{I}^{(\ell)}=\{\ell_1,\ell_2,\cdots,\ell_{n_0} \}$. 
Since $\{X_{\ell_1,t}\}_{t\ge 0}, \{X_{\ell_2,t}\}_{t\ge 0},\cdots,\{X_{\ell_{n_0},t}\}_{t\ge 0}$ are i.i.d copies of $\{X_{0,t}\}_{t\ge 0}$, the $\beta$-mixing coefficient of $$\{X_{\ell_1,1},X_{\ell_1,2},\cdots,X_{\ell_1,T},X_{\ell_2,1},X_{\ell_2,2},\cdots,X_{\ell_2,T},\cdots,X_{\ell_{n_0},1},X_{\ell_{n_0},2},\cdots,X_{\ell_{n_0},T}\}$$
satisfies $\beta(t)=O(\rho^t)$ for any $t\ge 0$ as well. 

Let $\phi_{j,t,b}$ denote $\varphi^{*}_R(\mu_b|X_{j,t})-\widehat{\varphi}^{(-\ell)}_R(\mu_b|X_{j,t})$. By (C2), 
we have
\begin{eqnarray}\label{eq1}
\max_{j,t,b}\Mean^{X_{j,t}} \phi_{j,t,b}^4\le 4\max_{b\in \{1,\cdots,B\}} \int_x \{\varphi^{*}_R(\mu_b|x)-\widehat{\varphi}^{(-\ell)}_R(\mu_b|x)\}^2\mathbb{F}(dx)\equiv \Delta,
\end{eqnarray}
where the expectation $\Mean^{X_{j,t}}$ is taken with respect to $X_{j,t}$. Notice that $\Delta$ is a random variable that depends on $\{\mu_b,\nu_b\}_{1\le b\le B}$ and $\{X_{j,t}\}_{j\in \mathcal{I}^{(-\ell)},0\le t\le T}$. By \eqref{eq1}, we have
\begin{eqnarray*}
	\max_{j,t,b}\Mean^{X_{j,t}} (\phi_{j,t,b}^2-\Mean^{X_{j,t}} \phi_{j,t,b}^2)^2\le \Delta.
\end{eqnarray*}
Under the boundedness assumption, we have $|\phi_{j,t,b}|\le 2$ and hence $|\phi_{j,t,b}^2-\Mean^{X_{j,t}} \phi_{j,t,b}^2|\le 4$. 

By Theorem 4.2 of \citet{Chen2015}, we have for any integers $\tau\ge 0$ and $1<d<n_0T/2$ that
\begin{eqnarray*}
	\prob\left(\left.\left|\sum_{j\in \mathcal{I}^{(\ell)}}\sum_{t=1}^T (\phi_{j,t,b}^2-\Mean^{X_{0,0}} \phi_{0,0,b}^2) \right|\ge 6\tau\right|\Delta \right)\le \frac{n_0T}{d}\beta(d)+\prob\left( \left.\left|\sum_{(j,t)\in \mathcal{I}_r}(\phi_{j,t,b}^2-\Mean^{X_{0,0}} \phi_{0,0,b}^2) \right|_2\ge \tau\right|\Delta \right)\\
	+4\exp\left( -\frac{\tau^2/2}{n_0Td\Delta+4d\tau/3} \right),
\end{eqnarray*}
where $\mathcal{I}_r$ denotes the last $n_0T-d\floor{n_0T/d}$ elements in the list
\begin{eqnarray}\label{list}
\{(\ell_1,1),(\ell_1,2),\cdots,(\ell_1,T),(\ell_2,1),(\ell_2,2),\cdots,(\ell_2,T),\cdots,(\ell_{n_0},1),(\ell_{n_0},2),\cdots,(\ell_{n_0},T)\},
\end{eqnarray} 
and $\floor{z}$ denote the largest integer that is smaller than or equal to $z$ for any $z$. Suppose $\tau\ge 4d$. Notice that $|\mathcal{I}_r|\le d$. It follows that
\begin{eqnarray*}
	\prob\left( \left.\left|\sum_{(j,t)\in \mathcal{I}_r}(\phi_{j,t,b}^2-\Mean^{X_{0,0}} \phi_{0,0,b}^2) \right|_2\ge \tau \right|\Delta\right)=0.
\end{eqnarray*}
Notice that $\beta(t)=O(\rho^t)$. Set $d=-(c^*+3)\log (n_0 T)/\log \rho$, we obtain $n_0T\beta(d)/d=O(n_0^{-2}T^{-2}B^{-1})=O(B^{-1}Q^{-1}n^{-2}T^{-2})$, since $Q\le T$, $B=O((nT)^{c_*})$ and $n_0=n/\mathbb{L}$. Here, the big-$O$ notation is uniform in $b\in \{1,\cdots,B\}$ and $q\in \{0,\cdots,Q\}$. Set $\tau=\max\{3\sqrt{\Delta n_0Td \log (Bn_0T)}, 11d\log(Bn_0 T)\}$, we obtain that
\begin{eqnarray*}
	\frac{\tau^2}{4}\ge 2n_0Td\Delta\log (BTn_0)\,\,\,\,\hbox{and}\,\,\,\,\frac{\tau^2}{4}\ge  8d\tau\log (BTn_0)/3\,\,\,\,\hbox{and}\,\,\,\,\tau \ge 4d,
\end{eqnarray*}
as either $n\to \infty$ or $T\to \infty$. It follows that $\tau^2/(2n_0Td\Delta+8d\tau/3)\ge 2\log (Bn_0T)$ and hence
\begin{eqnarray*}
	\max_{b\in \{1,\cdots,B\}}\max_{q\in \{0,\cdots,Q\}}\prob\left(\left.\left|\sum_{j\in \mathcal{I}^{(\ell)}}\sum_{t=1}^T (\phi_{j,t,b}^2-\Mean^{X_{0,0}} \phi_{0,0,b}^2) \right|\ge 6\tau\right|\Delta \right)=O(B^{-1}Q^{-1}n^{-1}T^{-1}).
\end{eqnarray*}
By Bonferroni's inequality, we obtain
\begin{eqnarray*}
	\prob\left(\left.\max_{b\in \{1,\cdots,B\}}\max_{q\in \{0,\cdots,Q\}}\left|\sum_{j\in \mathcal{I}^{(\ell)}}\sum_{t=1}^T (\phi_{j,t,b}^2-\Mean^{X_{0,0}} \phi_{0,0,b}^2) \right|\ge 6\tau\right|\Delta \right)=O(n^{-1}T^{-1}).
\end{eqnarray*}
Thus, with probability $1-O(n^{-1}T^{-1})$, we have
\begin{eqnarray}\label{eq2}
\max_{b\in \{1,\cdots,B\}}\max_{q\in \{0,\cdots,Q\}}\left|\sum_{j\in \mathcal{I}^{(\ell)}}\sum_{t=1}^T (\phi_{j,t,b}^2-\Mean^{X_{0,0}} \phi_{0,0,b}^2) \right|=O(\sqrt{\Delta n_0T}\log (Bn_0T),\log^2 (Bn_0T)).
\end{eqnarray}
Under the given conditions on $Q$, we have $T-q-1$ is proportional to $T$ for any $q\le Q$. Combining (C4) and the condition on $B$ with \eqref{eq2} yields \eqref{eqstep1eq4}. 

\textbf{Proof of \eqref{eqstep1eq1} with $m=2$: }
Similar to the proof of \eqref{eqstep1eq2}, it suffices to show $\max_{q,b} \sqrt{n(T-q-1)}|R_{2,\ell}(q,\mu_b,\nu_b)|=o_p(\log^{-1/2} (nT))$, or $\max_{q,b} \sqrt{n(T-q-1)}|R_{2,\ell}^{(r)}(q,\mu_b,\nu_b)|=o_p(\log^{-1/2} (nT))$ for any $\ell=1,\cdots,\mathbb{L}$ and $r=1,2,3,4$ where
\begin{eqnarray*}
	R_{2,\ell}(q,\mu,\nu)=\frac{1}{n(T-q-1)} \sum_{j\in \mathcal{I}^{(\ell)}} \sum_{t=1}^{T-q-1}\{
	\exp(i\mu^\top S_{j,t+q+1})
	-\varphi^{*}(\mu|X_{j,t+q}) \}\{\psi^*(\nu|X_{j,t})-\widehat{\psi}^{(-\ell)}(\nu|X_{j,t})\},\\
	R_{2,\ell}^{(1)}(q,\mu,\nu)=\frac{1}{n(T-q-1)} \sum_{j\in \mathcal{I}^{(\ell)}} \sum_{t=1}^{T-q-1}\{
	\cos(\mu^\top S_{j,t+q+1})
	-\varphi^{*}_R(\mu|X_{j,t+q}) \}\{\psi^*_R(\nu|X_{j,t})-\widehat{\psi}^{(-\ell)}_R(\nu|X_{j,t})\},\\
	R_{2,\ell}^{(2)}(q,\mu,\nu)=\frac{1}{n(T-q-1)} \sum_{j\in \mathcal{I}^{(\ell)}} \sum_{t=1}^{T-q-1}\{
	\sin(\mu^\top S_{j,t+q+1})
	-\varphi^{*}_I(\mu|X_{j,t+q}) \}\{\psi^*_I(\nu|X_{j,t})-\widehat{\psi}^{(-\ell)}_I(\nu|X_{j,t})\},\\
	R_{2,\ell}^{(3)}(q,\mu,\nu)=\frac{1}{n(T-q-1)} \sum_{j\in \mathcal{I}^{(\ell)}} \sum_{t=1}^{T-q-1}\{
	\cos(\mu^\top S_{j,t+q+1})
	-\varphi^{*}_R(\mu|X_{j,t+q}) \}\{\psi^*_I(\nu|X_{j,t})-\widehat{\psi}^{(-\ell)}_I(\nu|X_{j,t})\},\\
	R_{2,\ell}^{(4)}(q,\mu,\nu)=\frac{1}{n(T-q-1)} \sum_{j\in \mathcal{I}^{(\ell)}} \sum_{t=1}^{T-q-1}\{
	\sin(\mu^\top S_{j,t+q+1})
	-\varphi^{*}_I(\mu|X_{j,t+q}) \}\{\psi^*_R(\nu|X_{j,t})-\widehat{\psi}^{(-\ell)}_R(\nu|X_{j,t})\}.
\end{eqnarray*}
In the following, we only show $\max_{q,b} \sqrt{n(T-q-1)}|R_{2,\ell}^{(1)}(q,\mu_b,\nu_b)|=o_p(\log^{-1/2} (nT))$ to save space. 

Define the list
\begin{eqnarray*}
	\{(\ell_1,1),(\ell_1,2),\cdots,(\ell_1,T-q),(\ell_2,1),(\ell_2,2),\cdots,(\ell_2,T-q)\cdots,(\ell_{n_0},1),(\ell_{n_0},2),\cdots,(\ell_{n_0},T-q) \}.
\end{eqnarray*}
For any $1\le g\le n_0(T-q)$, denote by $(n_g,T_g)$ the $g$-th element in the list. Let $\mathcal{F}^{(0)}_q=\{X_{\ell_1,1},X_{\ell_1,2},\cdots,X_{\ell_1,1+q}\}\cup \{X_{j,t}:0\le t\le T,j\in \mathcal{I}^{(-\ell)} \}\cup \{\mu_1,\cdots,\mu_B,\nu_1,\cdots,\nu_B\}$. Then we recursively define $\mathcal{F}^{(g)}_q$ as
\begin{eqnarray*}
	\mathcal{F}^{(g)}_q=\left\{\begin{array}{ll}
		\mathcal{F}^{(g-1)}_q\cup \{X_{n_g,t_g+q+1}\}, & \hbox{if}~g=1~\hbox{or}~n_g=n_{g-1};\\
		\mathcal{F}^{(g-1)}_q\cup \{X_{n_{g-1},T},X_{n_g,1},X_{n_g,2},\cdots,X_{n_g,1+q}\}, & \hbox{otherwise}.
	\end{array}
	\right.
\end{eqnarray*}
Let $\phi_{g,q,b}^*=\{\cos(\mu_b^\top S_{n_g,t_g+q+1})
-\varphi^{*}_R(\mu_b|X_{n_g,t_g+q}) \}\{\psi^*_R(\nu_b|X_{n_g,t_g})-\widehat{\psi}^{(-\ell)}_R(\nu_b|X_{n_g,t_g})\}$.
Under MA, $R_{2,\ell}^{(1)}(q,\mu_b,\nu_b)$ can be rewritten as $\{n(T-q-1)\}^{-1}\sum_{g=1}^{n_0(T-q)} \phi^*_{g,q,b}$ and forms a sum of martingale difference sequence with respect to the filtration $\{\sigma(\mathcal{F}^{(g)}_q):g\ge 0\}$ where $\sigma(\mathcal{F}^{(g)}_q)$ denotes the $\sigma$-algebra generated by variables in $\mathcal{F}^{(g)}_q$. In the following, we apply concentration inequalities for martingales to bound $\max_{q,b} |R_{2,\ell}^{(1)}(q,\mu_b,\nu_b)|$. 

Under the boundedness condition, we have $|\phi_{g,q,b}^*|^2\le 4 \{\psi^*_R(\nu_b|X_{n_g,t_g})-\widehat{\psi}^{(-\ell)}_R(\nu_b|X_{n_g,t_g})\}^2$. In addition, we have by MA that
\begin{eqnarray*}
	\Mean\{(\phi_{g+1,q,b}^{*})^2|\sigma(\mathcal{F}^{(g)}_q)\}=\Mean [\{\cos(\mu_b^\top S_{n_g,t_g+q+1})
	-\varphi^{*}_R(\mu_b|X_{n_g,t_g+q}) \}^2|X_{n_g,t_g+q}]\\ \times \{\psi^*_R(\nu_b|X_{n_g,t_g})-\widehat{\psi}^{(-\ell)}_R(\nu_b|X_{n_g,t_g})\}^2
	\le 4\{\psi^*_R(\nu_b|X_{n_g,t_g})-\widehat{\psi}^{(-\ell)}_R(\nu_b|X_{n_g,t_g})\}^2. 
\end{eqnarray*}
It follows from Theorem 2.1 of \citet{Bercu2008} that
\begin{eqnarray*}
	\prob\left(\left|\sum_{g=1}^{n_0(T-q)} \phi^*_{g,q,b}\right|\ge\tau, \sum_{g=1}^{n_0(T-q)} 4\{\psi^*_R(\nu_b|X_{n_g,t_g})-\widehat{\psi}^{(-\ell)}_R(\nu_b|X_{n_g,t_g})\}^2\le y \right)\le 2\exp\left(-\frac{\tau^2}{2y}\right),\,\,\,\,\forall y,\tau,
\end{eqnarray*}
and hence
\begin{eqnarray*}
	\prob\left(\left|\sum_{g=1}^{n_0(T-q)} \phi^*_{g,q,b}\right|\ge\tau, \max_{b\in\{1,\cdots,B\}}\sum_{j\in \mathcal{I}^{(\ell)}} \sum_{t=1}^T \{\psi^*_R(\nu_b|X_{j,t})-\widehat{\psi}^{(-\ell)}_R(\nu_b|X_{j,t})\}^2\le \frac{y}{4} \right)\le 2\exp\left(-\frac{\tau^2}{2y}\right),\,\,\,\,\forall y,\tau,
\end{eqnarray*}
By Bonferroni's inequality, we obtain
\begin{eqnarray*}
	\prob\left(\max_{\substack{q\in \{0,\cdots,Q\} \\ b\in \{1,\cdots,B\} }}\left|\sum_{g=1}^{n_0(T-q)} \phi^*_{g,q,b}\right|\ge\tau, \max_{b\in\{1,\cdots,B\}}\sum_{j\in \mathcal{I}^{(\ell)}} \sum_{t=1}^T \{\psi^*_R(\nu_b|X_{j,t})-\widehat{\psi}^{(-\ell)}_R(\nu_b|X_{j,t})\}^2\le \frac{y}{4} \right)\le 2BQ\exp\left(-\frac{\tau^2}{2y}\right),
\end{eqnarray*}
for any $y,\tau$. Set $y=4\varepsilon \sqrt{nT}$, we obtain
\begin{eqnarray*}
	\prob\left(\max_{\substack{q\in \{0,\cdots,Q\} \\ b\in \{1,\cdots,B\} }}\left|\sum_{g=1}^{n_0(T-q)} \phi^*_{g,q,b}\right|\ge\tau, \max_{b\in\{1,\cdots,B\}}\sum_{j\in \mathcal{I}^{(\ell)}} \sum_{t=1}^T \{\psi^*_R(\nu_b|X_{j,t})-\widehat{\psi}^{(-\ell)}_R(\nu_b|X_{j,t})\}^2\le \sqrt{nT} \right)\\\le 2BQ\exp\left(-\frac{\tau^2}{2\sqrt{nT}}\right),
\end{eqnarray*}
It follows from \eqref{eqstep1eq5} that 
\begin{eqnarray}\label{eq3}
\prob\left(\max_{\substack{q\in \{0,\cdots,Q\} \\ b\in \{1,\cdots,B\} }}\left|\sum_{g=1}^{n_0(T-q)} \phi^*_{g,q,b}\right|\ge\tau\right)\le 2BQ\exp\left(-\frac{\tau^2}{2 \sqrt{nT}}\right)+o(1).
\end{eqnarray}
Set $\tau=(nT)^{1/4}\sqrt{2\log (BQnT)}$, the right-hand-side (RHS) of \eqref{eq3} is $o(1)$. Under the given conditions on $B$ and $Q$, we obtain  $\max_{q,b} \sqrt{n(T-q-1)}|R_{2,\ell}^{(1)}(q,\mu_b,\nu_b)|=o_p(\log^{-1/2} (nT))$. 

\subsubsection{Step 2}
For any $j\in \mathcal{I}^{(\mathbb{\ell})}$ and $0< t<T-q$, define vectors $\lambda_{R,q,j,t}^*,\lambda_{I,q,j,t}^*\in \mathbb{R}^{\mathbb{B}}$ such that the $b$-th element of $\lambda_{R,q,j,t}^*,\lambda_{I,q,j,t}^*$ correspond to the real and imaginary part of 
\begin{eqnarray*}
	\frac{1}{\sqrt{n(T-q-1)}}\{\exp(i\mu_b^\top S_{j,t+q+1})
	-\varphi^{*}(\mu_b|X_{j,t+q}) \}\{\exp(i\nu_b^\top X_{j,t-1})-\psi^{*}(\nu_b|X_{j,t})\},
\end{eqnarray*}
respectively. Let $\lambda^*_{q,j,t}$ denote the (2B)-dimensional vector $(\lambda_{R,q,j,t}^{*\top},\lambda_{I,q,j,t}^{*\top})^\top$. In addition, we define a (2B(Q+1))-dimensional vector $\lambda_{j,t}^*$ as
$(\lambda_{0,j,t}^{*\top}, \lambda_{1,j,t-1}^{*\top}\mathbb{I}(t>1), \cdots,\lambda_{1,j,t-Q}^{*\top}\mathbb{I}(t>Q))^\top$.
Define the list
\begin{eqnarray}
(1,1), (1,2),\cdots,(1,T-1),(2,1), (2,2),\cdots,(2,T-1),\cdots,(n,1),(n,2),\cdots,(n,T-1). 
\end{eqnarray}
For any $1\le g\le n(T-1)$, let $(n_g,t_g)$ be the $g$-th element in the list. Let $\mathcal{F}^{(0)}=\{X_{1,0}\}\cup \{\mu_1,\cdots,\mu_B,\nu_1,\cdots,\nu_B\}$ and recursively define $\mathcal{F}^{(g)}$ as
\begin{eqnarray*}
	\mathcal{F}^{(g)}=\left\{\begin{array}{ll}
		\mathcal{F}^{(g-1)}\cup \{X_{n_g,t_g}\}, & \hbox{if}~g=1~\hbox{or}~n_g=n_{g-1};\\
		\mathcal{F}^{(g-1)}\cup \{X_{n_{g-1},T},X_{n_g,0}\}, & \hbox{otherwise}.
	\end{array}
	\right.
\end{eqnarray*}
The high-dimensional vector $M_{n,T}=\sum_{g=1}^{n(T-1)} \lambda_{n_g,t_g}^*$ forms a sum of martingale difference sequence with respect to the filtration $\{\sigma(\mathcal{F}^{(g)}):g\ge 0 \}$. Notice that $S^*=\|\sum_{g=1}^{n(T-1)} \lambda_{n_g,t_g}^*\|_{\infty}$. In this step, we apply the high-dimensional martingale central limit theorem developed by \citet{belloni2018} to establish the limiting distribution of $S^*$. 

For $1\le g\le n(T-1)$, let
\begin{eqnarray*}
	\Sigma_g=\sum_{g=1}^{n(T-1)} \Mean \left(\left.\lambda_{n_g,t_g}^* \lambda_{n_g,t_g}^{*\top}\right|\mathcal{F}^{(g-1)} \right).
\end{eqnarray*}
Let $V^*=\sum_{g=1}^{n(T-1)} \Sigma_g$. Using similar arguments in proving \eqref{eq2}, we can show $\|V^*-V_0\|_{\infty,\infty}=O((nT)^{-1/2}\log (BnT))+O((nT)^{-1}\log^2 (BnT))$, with probability $1-O(n^{-1}T^{-1})$, where $V_0=\Mean V^*$. Under the given conditions on $B$, we have $\|V^*-V_0\|_{\infty,\infty}\le \kappa_{B,n,T}$ for some $\kappa_{B,n,T}=O((nT)^{-1/2}\log (nT))$, with probability $1-O(n^{-1}T^{-1})$. 

In addition, under the boundedness assumption in (C4), all the elements in $V^*$ and $V_0$ are uniformly bounded by some constants. It follows that $$\Mean \|V^*-V_0\|_{\infty,\infty}\le \kappa_{B,n,T}+\prob (\|V^*-V_0\|_{\infty,\infty}>\kappa_{B,n,T})=O((nT)^{-1/2}\log (nT)).$$ 
By Theorem 3.1 of  \citet{belloni2018}, we have for any Borel set $\mathcal{R}$ and any $\delta>0$ that
\begin{eqnarray}\label{importantassertion}
&&\prob(S^*\in \mathcal{R})\le \prob (\|N(0,V_0)\|_{\infty}\in \mathcal{R}^{C\delta})|\\\nonumber&\le& C\left(\frac{1}{nT}+\frac{\log (BnT)\log (BQ)}{\delta^2\sqrt{nT}}+\frac{\log^3 (BQ)}{\delta^3 \sqrt{nT}}+\frac{\log^3(BQ)}{\delta^3}\sum_{g=1}^{n(T-1)}\Mean \|\eta_g\|_{\infty}^3 \right),
\end{eqnarray}
for some constant $C>0$. 

Under the boundedness assumption in (C4), the absolute value of each element in $\Sigma_g$ is uniformly bounded by $16 (n(T-q-1))^{-1}=O(n^{-1}T^{-1})$. With some calculations, we can show that $\sum_{g=1}^{n(T-1)}\Mean \|\eta_g\|_{\infty}^3=O((nT)^{-1/2}\log^{3/2} (BQ))$. In addition, we have $Q=O(T)$ and $B=O((nT)^{c_*})$. Combining these together with \eqref{importantassertion} yields 
\begin{eqnarray}\label{importantassertion2}
\prob(S^*\in \mathcal{R})\le \prob (\|N(0,V_0)\|_{\infty}\in \mathcal{R}^{C\delta})|+ O(1)\left(\frac{1}{nT}+\frac{
	\log^2 (nT)}{\delta^2\sqrt{nT}}+\frac{\log^{9/2}(nT)}{\delta^3\sqrt{nT}} \right),
\end{eqnarray}
where $O(1)$ denotes some positive constant. 

Set $\mathcal{R}=(z,+\infty)$ and $\delta=\varepsilon \log^{-1/2} (nT)/C$, we obtain
\begin{eqnarray*}
	\prob(S^*\le z)\ge \prob(\|N(0,V_0)\|_{\infty}\le z-\varepsilon \log^{-1/2} (nT))-o(1).
\end{eqnarray*}
Set $\mathcal{R}=(-\infty,z]$, we can similarly show
\begin{eqnarray*}
	\prob(S^*\le z)\le \prob(\|N(0,V_0)\|_{\infty}\le z+\varepsilon \log^{-1/2} (nT))+o(1).
\end{eqnarray*} 
This completes the proof of Step 2. 

\subsubsection{Step 3}
We break the proof into two parts. In Part 1, we show $V_0$ is a block diagonal matrix. Specifically, let $V_{0,q_1,q_2}$ denote the $(2B)\times (2B)$ submatrix of $V_0$ formed by rows in $\{2q_1B+1,2q_1B+2,\cdots,2(q_1+1)B\}$ and columns in  $\{2q_2B+1,2q_2B+2,\cdots,2(q_2+1)B\}$. For any $q_1\neq q_2$, we show $V_{0,q_1,q_2}=O_{(2B)\times (2B)}$. 

Let $\Sigma^{(q)}$ denote $V_{0,q,q}$. In Part 2, we provide an upper bound for $\max_{q\in \{0,\cdots,Q\}}\|\Sigma^{(q)}-\widehat{\Sigma}^{(q)}\|_{\infty,\infty}$. Let $\widehat{V}$ be a block diagonal matrix where the main diagonal blocks are given by $\widehat{\Sigma}^{(0)},\widehat{\Sigma}^{(1)},\cdots,\widehat{\Sigma}^{(Q)}$, we obtain $\|V_0-\widehat{V}\|_{\infty,\infty}$

\textbf{Part 1: }Let $\lambda_{R,q,j,t,b}^*$ and $\lambda_{I,q,j,t,b}^*$ denote the $b$-th element of $\lambda_{R,q,j,t}^*$ and $\lambda_{I,q,j,t}^*$, respectively. Each element in $V_{0,q_1,q_2}$ equals $\Mean (\sum_{j,t} \lambda_{Z_1,q_1,j,t,b_1}^*) (\sum_{j,t} \lambda_{Z_2,q_2,j,t,b_2}^*)$ for some $b_1,b_2\in \{1,\cdots,B\}$ and $Z_1,Z_2\in \{R,I\}$. In the following, we show 
\begin{eqnarray*}
	\Mean \left(\sum_{j,t} \lambda_{R,q_1,j,t,b_1}^*\right) \left(\sum_{j,t} \lambda_{R,q_2,j,t,b_2}^*\right)=0,\,\,\,\,\,\,\,\,\forall q_1\neq q_2.
\end{eqnarray*}
Similarly, one can show $\Mean (\sum_{j,t} \lambda_{R,q_1,j,t,b_1}^*) (\sum_{j,t} \lambda_{I,q_2,j,t,b_2}^*)=0$ and $\Mean (\sum_{j,t} \lambda_{I,q_1,j,t,b_1}^*) (\sum_{j,t} \lambda_{I,q_2,j,t,b_2}^*)=0$ for any $q_1\neq q_2$. This completes the proof for Part 1. 

Since observations in different trajectories are i.i.d, it suffices to show
\begin{eqnarray*}
	\sum_j \Mean \left(\sum_{t} \lambda_{R,q_1,j,t,b_1}^*\right) \left(\sum_{t} \lambda_{R,q_2,j,t,b_2}^*\right)=0,\,\,\,\,\,\,\,\,\forall q_1\neq q_2,
\end{eqnarray*}
or equivalently,
\begin{eqnarray}\label{eq4.5}
\Mean \left(\sum_{t} \lambda_{R,q_1,0,t,b_1}^*\right) \left(\sum_{t} \lambda_{R,q_2,0,t,b_2}^*\right)=0,\,\,\,\,\,\,\,\,\forall q_1\neq q_2,
\end{eqnarray}
By definition, we have
\begin{eqnarray*}
	\lambda_{R,q,0,t,b}^*=\frac{1}{\sqrt{n(T-q-1)}}\{\cos(\mu_b^\top S_{0,t+q+1})
	-\varphi_R^{*}(\mu_b|X_{0,t+q}) \}\{\cos(\nu_b^\top X_{0,t-1})-\psi_R^{*}(\nu_b|X_{0,t})\}.
\end{eqnarray*}
Since $q_1\neq q_2$, for any $t_1,t_2$, we have either $t_1+q_1\neq t_2+q_2$ or $t_1\neq t_2$. Suppose $t_1+q_1>t_2+q_2$. Under MA, we have 
\begin{eqnarray*}
	\Mean [\{\cos(\mu_b^\top S_{0,t_1+q_1+1})
	-\varphi_R^{*}(\mu_b|X_{0,t_1+q_1})\}|\{X_{0,j}\}_{j\le t_1+q_1}]=0, \,\,\,\,\,\,\,\,\forall b,
\end{eqnarray*}
and hence 
\begin{eqnarray}\label{eq4}
\Mean \lambda_{R,q_1,0,t_1,b_1}^* \lambda_{R,q_2,0,t_2,b_2}^*=0,\,\,\,\,\,\,\,\,\forall b_1,b_2.
\end{eqnarray}
Similarly, when $t_1+q_1<t_2+q_2$, we can show \eqref{eq4} holds as well. 

Suppose $t_1<t_2$, under (C1) and $H_0$, we have 
\begin{eqnarray*}
	\Mean [\{\cos(\nu_b^\top X_{0,t_1-1})
	-\varphi_R^{*}(\nu_b|X_{0,t_1})\}|\{X_{0,j}\}_{j\ge t_1}]=0, \,\,\,\,\,\,\,\,\forall b,
\end{eqnarray*}
and hence \eqref{eq4} holds. Similarly, when $t_1>t_2$, we can show \eqref{eq4} holds as well. This yields \eqref{eq4.5}. 

\textbf{Part 2: }For any $q\in \{0,\cdots,Q\}$, we can represent $\widehat{\Sigma}^{(q)}-\Sigma^{(q)}$ by
\begin{eqnarray}\label{eq5}
\sum_{\ell=1}^{\mathbb{L}} \sum_{j\in \mathcal{I}^{(\ell)}}\sum_{t=1}^{T-q-1} \frac{(\lambda_{R,q,j,t}^\top,\lambda_{I,q,j,t}^\top)^\top  (\lambda_{R,q,j,t}^\top,\lambda_{I,q,j,t}^\top)-(\lambda_{R,q,j,t}^{*\top},\lambda_{I,q,j,t}^{*\top})^\top  (\lambda_{R,q,j,t}^{*\top},\lambda_{I,q,j,t}^{*\top})}{n(T-q-1)}.
\end{eqnarray}
Using similar arguments in Step 1 of the proof, we can show with probability tending to $1$ that the absolute value of each element in \eqref{eq5} is upper bounded by $c_0^*(nT)^{-c^{**}}$ for any $q\in \{0,\cdots,Q\}$ and some positive constants $c_0,c^*>0$. Thus we obtain $\max_{q\in \{0,\cdots,Q\}} \|\widehat{\Sigma}^{(q)}-\Sigma^{(q)}\|_{\infty,\infty}=O((nT)^{-c^{**}})$, with probability tending to $1$. The proof is hence completed.

\end{document}